\documentclass{article} 
\usepackage{main,times}

\usepackage{amsmath,amsfonts,bm}

\def\eqref#1{equation~\ref{#1}}

\def\1{\bm{1}}

\DeclareMathAlphabet{\mathsfit}{\encodingdefault}{\sfdefault}{m}{sl}
\SetMathAlphabet{\mathsfit}{bold}{\encodingdefault}{\sfdefault}{bx}{n}

\usepackage[hidelinks]{hyperref}
\usepackage{url}

\usepackage{booktabs}
\usepackage{enumitem}
\usepackage{graphicx} 
\usepackage{caption}

\usepackage{multirow}
\usepackage{algorithm,algorithmicx,algpseudocode}
\usepackage{listings}

\title{MolChord: Structure-Sequence Alignment for Protein-Guided Drug Design}

\author{Wei Zhang\textsuperscript{1}\thanks{Equal contribution.} ,\;
Zekun Guo\textsuperscript{2}\footnotemark[1] ,\;
Yingce Xia\textsuperscript{2}\footnotemark[1]\hspace{0.4em}\thanks{Corresponding author.} ,\;
Peiran Jin\textsuperscript{2}\footnotemark[1]\hspace{0.4em}\footnotemark[2] ,\;
Shufang Xie\textsuperscript{2},\\
\textbf{Tao Qin\textsuperscript{2},\;
Xiang{-}Yang Li\textsuperscript{1}}\\
\textsuperscript{1}School of Computer Science and Technology, University of Science and Technology of China\\
\textsuperscript{2}Zhongguancun Academy\\
\texttt{weizhang\_cs@mail.ustc.edu.cn};\;\;\texttt{xiangyangli@ustc.edu.cn};\\
\texttt{\{guozekun,xiayingce,jinpeiran,xieshufang,qintao\}@bjzgca.edu.cn}
}

\newcommand{\ours}{\textsc{MolChord}}
\iclrfinalcopy 
\begin{document}
\maketitle

\begin{abstract}
Structure-based drug design (SBDD), which maps target proteins to candidate molecular ligands, is a fundamental task in drug discovery. Effectively aligning protein structural representations with molecular representations, and ensuring alignment between generated drugs and their pharmacological properties, remains a critical challenge. To address these challenges, we propose \ours{}, which integrates two key techniques: (1) to align protein and molecule structures with their textual descriptions and sequential representations (e.g., FASTA for proteins and SMILES for molecules), we leverage NatureLM, an autoregressive model unifying text, small molecules, and proteins, as the molecule generator, alongside a diffusion-based structure encoder; and (2) to guide molecules toward desired properties, we curate a property-aware dataset by integrating preference data and refine the alignment process using Direct Preference Optimization (DPO). Experimental results on CrossDocked2020 demonstrate that our approach achieves state-of-the-art performance on key evaluation metrics, highlighting its potential as a practical tool for SBDD.

\end{abstract}

\section{Introduction}

Drug discovery is a long and costly process, often spanning over a decade and requiring billions of dollars in investment~\citep{paul2010improve, dimasi2016innovation}. The chemical space is estimated to contain up to $10^{60}$ synthetically accessible molecules~\citep{polishchuk2013estimation}, making it infeasible explore all possibilities. Structure-based drug design (SBDD) has emerged as a transformative approach in drug discovery~\citep{anderson2003process, batool2019structure, schneider2020rethinking}, leveraging the structure of biological targets to rationally design drug compounds using computational techniques like molecular docking. Recent advances in artificial intelligence (AI) have further enhanced SBDD~\citep{luo20213d, peng2022pocket2mol, guan20233d}, with typical frameworks employing protein encoders to transform protein structures into high-dimensional representations and generators to map these representations back into the chemical space~\citep{wu2024tamgen, feng2024generation}, either as 3D molecular structures or chemical descriptors. These advancements significantly improve the efficiency and accuracy of drug design.

Despite these advancements, aligning protein representations with molecular representations remains a challenge for AI-based SBDD, mainly due to the limited number of high-quality protein–ligand pairs~\citep{feng2023protein, gao2023profsa}. Furthermore, ensuring that generated compounds are aligned with desired drug properties presents another critical issue. However, generating large-scale, high-quality protein–ligand data is prohibitively expensive and time-consuming~\citep{davies2006streamlining, nakata2023end}. Instead of solely relying on building more protein–ligand datasets with structural information, we propose exploring novel approaches to improve the alignment between structure encoders and chemical generators.

A promising trend in research is the development of unified scientific entity generators, such as MolXPT~\citep{liu2023molxpt} (text, small molecule), LucaOne~\citep{he2025generalized} (protein, DNA, RNA), and NatureLM~\citep{xia2025naturelm} (text, molecule, DNA, RNA, protein, material), which are designed to jointly model diverse biological and chemical sequences within a unified representational space. By adopting such a unified generator in AI-based SBDD models, alignment between structure encoders and molecule generators can be enhanced through tasks like protein-to-text and protein-to-FASTA transformations, whose data are substantially larger in scale compared to protein–ligand pairs. These tasks facilitate more effective alignment by enabling encoders and generators to learn across multiple modalities.


In this work, we introduce \ours{}, a four-billion-parameter framework with enhanced alignment between the structure encoder and sequence generator. The structure encoder follows the FlexRibbon framework~\citep{zhu2025flexribbon}, a diffusion-based model pre-trained to capture geometric and structural features (residue-level for proteins and atom-level for molecules). For the generator, we implement a variant of NatureLM~\citep{xia2025naturelm}, an autoregressive sequence generator capable of handling protein FASTA sequences, molecular SMILES, and text representations. Our training process consists of three stages to achieve robust alignment. First, the structure encoder and sequence generator are connected via a lightweight adapter, pre-trained on five structure-to-sequence tasks: protein-to-FASTA, protein-to-text, molecule-to-SMILES, molecule-to-text, and complex-to-FASTA/SMILES. This pre-training establishes a shared representational space across proteins and molecules. Next, we perform supervised fine-tuning on pocket–ligand complexes to anchor the model with biological evidence. Finally, we apply Direct Preference Optimization (DPO) to a curated subset of CrossDocked2020~\citep{francoeur2020three}, which provides reliable preference signals and broad protein coverage. This curation enables reinforcement learning to improve binding affinity while maintaining validity, synthesizability, and diversity. Through this staged design, \ours{} achieves scalable and effective protein--ligand alignment, yielding a unified foundation model that advances the practicality of SBDD.

We systematically evaluate \ours{} on CrossDocked2020~\citep{francoeur2020three}, the widely used dataset for SBDD. \ours{} consistently outperforms strong baselines on affinity-related proxies while preserving synthesizability (SA), quantitative estimate of drug-likeness (QED), and scaffold diversity. The gains are more pronounced under limited paired supervision and on held-out targets, indicating robust cross-modal alignment rather than overfitting to heuristics. Ablations show that both the diffusion-pretrained structure encoder and DPO fine-tuning are necessary; removing either degrades the affinity–drug-likeness trade-off. These results validate our design choice of coupling diffusion-based encoding with autoregressive generation via a lightweight sequential/textual adapter.

Our contribution can be summarized as follows:
\begin{itemize}[leftmargin=*]
\item We propose \ours{}, a unified framework that leverages diffusion to capture protein structure and autoregression for SMILES generation, aligning protein, molecule, and text representations in target-aware molecular design.  
\item We curate a property-aware dataset for reinforcement learning and apply Direct Preference Optimization (DPO) to refine alignment, improving binding affinity while preserving other molecular properties.
\item Experimental results on CrossDocked2020 datasets demonstrate that \ours{} achieves state-of-the-art performance on key evaluation metrics, underscoring its potential as a practical tool for structure-based drug design.  
\end{itemize}

\section{Related works}

\paragraph{Structure-based Drug Design} Structure-based drug design aims to design ligands conditioned on protein structures or sequences. Early representative works include liGAN~\citep{ragoza2022generating}, which voxelizes protein–ligand complexes into atomic density grids within a conditional VAE framework, and GraphBP~\citep{liu2022generating}, which generates ligands through graph-based placement in 3D binding pockets. Building on these foundations, recent work can be broadly categorized into three families: diffusion-based, flow-based, and autoregressive approaches. Diffusion-based methods model protein–ligand distributions in continuous 3D space, including DiffSBDD~\citep{schneuing2024structure}, TargetDiff~\citep{guan20233d} with SE(3)-equivariant denoising, and DecompDiff~\citep{guan2023decompdiff}, which incorporates functional-region decomposition to improve validity and synthesizability. Flow-based approaches parameterize generation in continuous latent space, such as FlowSBDD~\citep{zhang2024rectified} and MolForm~\citep{huang2025molform}, which leverage rectified or multimodal flow matching for molecular design. Autoregressive (AR) models formulate ligand design as conditional sequence generation. Early examples include AR~\citep{luo20213d}, Pocket2Mol~\citep{peng2022pocket2mol}, and ResGen~\citep{zhang2023resgen}, which autoregressively generate ligands conditioned on binding pockets. Among them, ResGen leverages residue-level encoding, while Pocket2Mol operates at the atom level. More recent developments adopt tokenization of structural inputs: XYZ-Transformer~\citep{flam2023language} and BindGPT~\citep{zholus2025bindgpt} directly treat 3D coordinates as tokens for autoregressive modeling. In addition, several works incorporate an explicit structure encoder to enrich conditional signals, including TamGen~\citep{wu2024tamgen}, 3D-SMILES-GPT~\citep{wang20253dsmiles}, and Lingo3DMol~\citep{feng2024generation}. This line of work is most closely related to our approach, yet our method distinguishes itself by scaling model capacity and introducing principled cross-modal alignment.

\paragraph{Reinforcement Learning} Likelihood training is standard in generative modeling, yet often misaligned with user objectives, motivating reinforcement learning for alignment. In particular, reinforcement learning from human feedback (RLHF)~\citep{ziegler2019fine, ouyang2022training} has proven effective in steering LLM toward human intent. More recently, Direct Preference Optimization (DPO)~\citep{rafailov2023direct} has emerged as a lightweight alternative that bypasses explicit reward modeling by directly optimizing on preference pairs, achieving results comparable to RLHF while being simpler and more stable to train. Recently, several studies have explored reinforcement learning in structure-based drug design. BindGPT~\citep{zholus2025bindgpt} and 3DMolFormer~\citep{hu20253dmolformer} integrate RL objectives to enhance binding affinity, while DecompDPO~\citep{cheng2024decomposed} introduces a decomposition-based alignment scheme to better guide optimization. Other approaches have incorporated preference-based learning into SBDD: MolForm~\citep{huang2025molform} applies Direct Preference Optimization (DPO) to improve docking affinity, and AliDiff~\citep{gu2024aligning} proposes Exact Energy Preference Optimization (E$^2$PO) with additional regularization. Despite these advances, BindGPT, 3DMolFormer, and DecompDPO tend to improve affinity at the cost of molecular diversity, whereas preference-based approaches like MolForm and AliDiff remain heavily tied to docking scores, often degrading key properties such as QED and synthesizability. These limitations point to the need for higher-quality preference data and more principled optimization objectives.

\section{Method}

In this section, we present \ours{}, our framework for structure-based drug design. We begin with the problem definition in Section~\ref{3:problem}, and then describe the overall architecture in Section~\ref{3:architecture}. The training strategy is introduced in Section \ref{sec:training_strategy}.

\subsection{Problem Definition} 
\label{3:problem}

SBDD can be formulated as conditional molecule generation given a protein pocket. Let
\(P^{\text{prot}} = \{(\mathbf{x}_i^{\text{res}}, \mathbf{a}_i)\}_{i=1}^{N_{\text{res}}}\) denote a protein, where \(\mathbf{x}_i^{\text{res}} \in \mathbb{R}^3\) is the 3D coordinate of the $\alpha$-carbon atom of the $i$-th residue, and \(\mathbf{a}_i\) denotes residue-level annotations such as amino acid type, chain identity, and residue index. A binding pocket \(P^{\text{pock}} \subset P^{\text{prot}}\) is defined as the subset of residues surrounding the active site. The goal of SBDD is to generate a ligand $M$ that can bind to $P^{\text{pock}}$. In this work, we focus on designing compounds in chemical space and let $M$ denote the SMILES sequence \(M = (m_1, m_2, \ldots, m_{\vert M\vert})\) with $s_i$ representing the $i$-th token in the SMILES sequence.

\subsection{Model Architecture}
\label{3:architecture}

\begin{figure}[!t]
    \centering
    \includegraphics[width=0.95\linewidth]{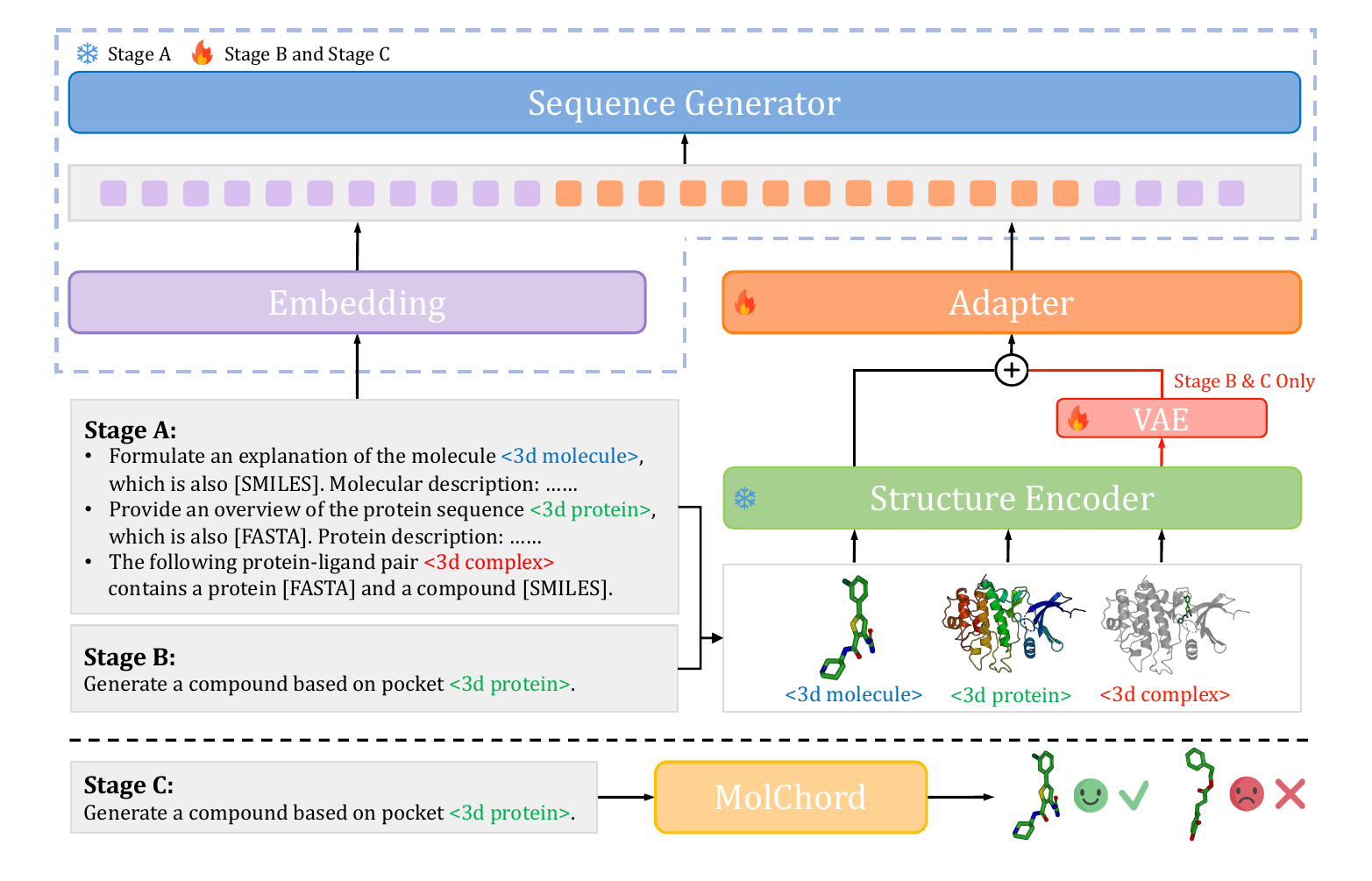}
    \caption{Overview of \ours{}. For each input, unmarked text tokens are embedded by the language model, while color-marked entities ($\langle\rm 3d\; molecule\rangle$, $\langle\rm 3d\; protein\rangle$, or $\langle\rm 3d\; complex\rangle$) are processed by the \texttt{Encoder}. In Stage~B, protein–ligand complexes are further processed through a VAE to perturb protein features, and only pocket features are injected into the language model. The bottom panel illustrates Stage~C, where Direct Preference Optimization (DPO) is applied.}
    \label{fig:framework}
\end{figure}

As illustrated in Figure~\ref{fig:framework}, the architecture consists of three main modules: a structure encoder (\texttt{Encoder}) that encodes structures of molecule, protein and complex; a sequence generator (\texttt{Generator}) responsible for generating SMILES and related sequences; an adapter (\texttt{Adapter}) with an auxiliary variational autoencoder (\texttt{VAE}) to align \texttt{Encoder} and \texttt{Generator}. Our model has 4.2B parameters in total. For each reference, denote the embedding layer of the \texttt{Generator} as \texttt{embed}, which maps discrete sequence tokens into hidden representations. 


\paragraph{Structure Encoder} The \texttt{Encoder}, pre-trained with a diffusion-based objective following FlexRibbon~\citep{zhu2025flexribbon}, is capable of processing protein, molecule, and protein–molecule complex structures within a single model. The input is defined as \(X = \{(\mathbf{x}_i, \mathbf{a}_i)\}_{i=1}^{N_{\text{tok}}}\), where $\mathbf{x}_i$ and $\mathbf{a}_i$ denote the coordinates and the annotation of the $i$-th element in $X$. Protein structures are represented at the residue level, molecular structures are represented at the atom level, and complex structures are represented with a combination of residues for the protein component and atoms for the molecular component. The architecture of the \texttt{Encoder} primarily follows the Elucidated Diffusion Model (EDM)~\citep{karras2022elucidating}, a variant of the Transformer architecture that incorporates geometric information. \texttt{Encoder} is pre-trained on AlphaFoldDB~\citep{varadi2024alphafold} and PDB~\citep{berman2000protein}. Additional details about pre-training setups can be found in Appendix~\ref{app:structure_encoder}. By using the \texttt{Encoder}, for each $(\mathbf{x}_i, \mathbf{a}_i)\in X$, we can obtain a contextual representation \texttt{Encoder}$(X)$.

\paragraph{Generator} Following \citet{xia2025naturelm}, the \texttt{Generator} is a language model pretrained on molecule SMILES, protein FASTA sequences, and textual annotations by using next token prediction. Further details are provided in Appendix~\ref{app:sequence_gen}. The pretraining of the \texttt{Encoder} and the \texttt{Generator} is conducted independently.

\paragraph{Align the \texttt{Encoder} and \texttt{Generator}} Given a 3D structure input $X$ and its corresponding annotation, we demonstrate how the \texttt{Encoder} and \texttt{Generator} are jointly utilized. Together, they form an interleaved sequence like:
\begin{equation}
I = (t_1,t_2,\cdots,t_m, (\mathbf{x}_1, \mathbf{a}_1), (\mathbf{x}_2, \mathbf{a}_2),\cdots,(\mathbf{x}_i, \mathbf{a}_{N_{\rm tok}}), t_{m+1},t_{m+2},\cdots,t_n),
\label{eq:interleaved_sequence}
\end{equation}
where $t_i$ represents tokens such as text, SMILES, or FASTA. 

For instance, see the first input of Stage~A in Figure~\ref{fig:framework}, where the prefix $(t_1,\ldots,t_m)$ corresponds to the text \emph{“Formulate an explanation of the molecule”}, the suffix $(t_{m+1},\ldots,t_n)$ corresponds to \emph{“which is also [SMILES]. Molecular description: …”}, and the placeholder $\langle\rm 3d\; molecule\rangle$ is expanded into $(\mathbf{x}_1,\mathbf{a}_1),\ldots,(\mathbf{x}_{N_{\rm tok}},\mathbf{a}_{N_{\rm tok}})$, which together constitute the 3D input $X$ of the molecule in $I$.

The 3D input $X$ in $I$ is first processed as 
\begin{equation}
\mathbf{U} = \texttt{Adapter}(\texttt{Encoder}(X)),
\label{eq:adapter-stagea}
\end{equation}
where each $(\mathbf{x}_i, \mathbf{a}_i)$ in $X$ is processed into $\mathbf{u}_i$, a high-dimensional representations and $U=(\mathbf{u}_1,\mathbf{u}_2,\cdots)$. The $t_i$ in $I$ is mapped by \texttt{embed} layer and obtain $e_i=\texttt{embed}(t_i)$. By this way, all elements in $I$ are mapped as 
\begin{equation}
I_{\rm emb} = \left(e_1, \cdots, e_m,\mathbf{u}_1,\cdots,\mathbf{u}_N, e_{m+1}, \cdots, e_{n}\right).
\label{eq:interleaved_embedding}
\end{equation}
The embedded sequence $I_{\rm emb}$ is then fed into the embedding layer of \texttt{Generator} to perform the generation task. This formulation unifies structural and textual tokens into a single embedding sequence, allowing the \texttt{Generator} to attend jointly over structural representations and symbolic annotations.

\subsection{Training strategy}
\label{sec:training_strategy}

We adopt a three-stage training strategy. In Stage A, we train only the parameters of the \texttt{Adapter} to align the \texttt{Encoder} with the \texttt{Generator}. In Stage B, we perform supervised fine-tuning on protein-ligand data to enhance the protein-to-ligand generation capability. Finally, in Stage C, we apply direct preference optimization (DPO) to align the model with key preferences essential for SBDD.

Denote the dataset of stage A as $\mathcal{D}_{\rm A}$, which consists of the following datasets for alignment: (i) 676K protein structures paired with FASTA sequences and functional annotations, collected from multiple sources including PDB~\citep{berman2000protein} and SwissProt~\citep{boutet2007uniprotkb}; (ii) 316K small molecules paired with SMILES and textual descriptions, collected from Uni-Mol~\citep{zhou2023uni}; and (iii) 94K protein–ligand complexes annotated with both 3D coordinates, obtained from PDB~\citep{berman2000protein}.  All datasets are processed into interleaved sequences (see Eqn. (\ref{eq:interleaved_sequence})).

For Stages B and C, we exclusively use protein–ligand complexes from CrossDocked2020~\citep{francoeur2020three}, which are subsequently divided into two disjoint datasets: $\mathcal{D}_{\rm B}$ and $\mathcal{D}_{\rm C}$. If a protein is associated with $>2$ molecules, it is assigned to $\mathcal{D}_{\rm B}$; otherwise it is assigned to $\mathcal{D}_{\rm C}$. The intuition behind this strategy is two-fold: (i) In large language model (LLM) training, it is typical to maintain disjoint datasets for supervised fine-tuning (SFT) and reinforcement learning (or decision preference optimization), as these stages have distinct objectives; (ii) for our task, if a protein pocket is associated with only one ligand, the \texttt{Generator} is less likely to produce diverse molecules, making it less effective for alignment purposes. Assigning such pairs to $\mathcal{D}_{\rm C}$ ensures a focus on alignment, while $\mathcal{D}_{\rm B}$ benefits from more diverse multi-ligand associations.

\paragraph{Stage A:} We freeze the \texttt{Encoder} and \texttt{Generator}, training only the \texttt{Adapter} that maps structural features to the embedding space of the \texttt{Generator}. This is achieved through next-token prediction:
\begin{equation}
\mathcal{L}_{\text{alignment}} =  - \frac{1}{\vert \mathcal{D}_A\vert}\sum_{I\in\mathcal{D}_A}\sum_{i=\texttt{fid}(I)}^{\vert I\vert}\log P(I_{i}|I_{<i}); 
\label{eq:align-loss}
\end{equation}
where $\texttt{fid}(I)$ denotes the first index following the 3D structure element (i.e, the index of $e_{m+1}$ in the $I$ of Eqn. (\ref{eq:interleaved_sequence})), and $\vert I \vert$ is the sequence length of $I$. 

\paragraph{Stage B:} The model is then fine-tuned on the protein-ligand dataset. We adopt a variational autoencoder (VAE)-based approach in this stage to increase the diversity of the generated molecules.
During training, a controlled noise term is injected into the \texttt{Adapter} as follows:
\begin{equation}
\begin{aligned}
& (\boldsymbol{\mu,\Sigma}) = \texttt{VAE}(\texttt{Encoder}(P^{\texttt{prot}}, M^{\texttt{ref}})), \\
& \mathbf{u} = \texttt{Adapter}\left(\texttt{Encoder}(P^{\texttt{prot}}) + \boldsymbol{\epsilon}\right).
\end{aligned}
\label{eq:adapter-stageb}
\end{equation}
In Eqn. (\ref{eq:adapter-stageb}), (i) \texttt{VAE} is a feed-forward layer that outputs the mean $\boldsymbol{\mu}$ and variance $\boldsymbol{\Sigma}$; (ii) $\boldsymbol{\epsilon}$ is sampled from the Gaussian distribution $\mathcal{N}(\boldsymbol{\mu},\boldsymbol{\Sigma})$. During inference, $\boldsymbol{\epsilon}$ is sampled from standard Gaussian distribution $\mathcal{N}(\bm{0}, \bm{I})$.

The output $\mathbf{u}$ is then used to construct a new interleaved sequence $I$ in Eqn. (\ref{eq:interleaved_embedding}). During Stage B, the \texttt{Encoder} and \texttt{Adapter} process the entire protein structure, while only the features corresponding to the binding pocket are injected into the \texttt{Generator}.

The overall training objective for Stage B is defined as:
\begin{equation}
\mathcal{L}_{\text{SFT}} = - \frac{1}{\vert\mathcal{D}_{\rm B}\vert} \sum_{I\in\mathcal{D}_{\rm B}}\sum_{i=\texttt{ind}(I)}^{\vert I\vert} \log P\left(I_{i}|I_{<i}\right) \;+\; 
\beta_{\text{vae}} \, D_{\mathrm{KL}}\!\left[p(\boldsymbol{\epsilon}) \Vert \mathcal{N}(\bm{0}, \bm{I})\right],
\label{eq:sft-loss}
\end{equation}
where \(\beta_{\text{vae}}>0\) is the hyperparameter.

\paragraph{Stage C:} The core aspect of DPO is constructing the preference data. For each pocket in $\mathcal{D}_{\rm C}$, we sample 100 candidate molecules using the checkpoint from Stage B with the lowest validation loss.
A pocket is retained for further processing if the diversity among these 100 candidates exceeds 0.8.
The diversity is measured as $1-\sum_{i=1}^{100}\sum_{j=i+1}^{100}\texttt{fingerprint\_similarity}(M_i, M_j)/Z$ where $Z$ is the normalization factor. By this way, about 1K protein pockets are selected, denoted as $\mathcal{D}_{\rm DPO}$. The reward for each sampled molecule $M$ is then defined as: 
\begin{equation}
R(M, P^{\text{pock}}) = -\left(S_{\text{Vina}}(M, P^{\text{pock}}) + \lambda \cdot \max(0,\; \#\texttt{fused\_ring}(M)-2)\right)
\end{equation}
where \(S_{\text{Vina}}\) is the docking score computed by AutoDock Vina (a lower docking score indicates better binding affinity), $\lambda$ denotes fused ring penalty, and \#\texttt{fused\_ring}$(M)$ represents the number of fused rings in molecule $M$ (a lower fused ring count may suggest that $M$ is easier to synthesize and have reduced toxicity). This quantity is strongly correlated with the molecule's quantitative estimate of drug-likeness (QED) and its synthetic accessibility. The molecules with the highest and lowest rewards are denoted as $M^+$ and $M^-$ respectively. The reward function is defined as follows:
\begin{equation}
\mathcal{L}_{\text{DPO}} = -\log \sigma \left( \beta_{\text{DPO}} \left[ 
\log \frac{\pi(M^+ \mid P^{\rm pock})}{\pi_{\text{ref}}(M^+ \mid P^{\rm pock})} 
- \log \frac{\pi(M^- \mid P^{\rm pock})}{\pi_{\text{ref}}(M^- \mid P^{\rm pock})} 
\right] \right),
\end{equation}
where \(\pi_{\text{ref}}\) is the frozen model from Stage B and \(\beta_{\text{DPO}}\) controls preference sharpness. Note that the variational encoder loss is also included in Stage C.

\section{Experiments}

\subsection{Experimental Setup}

\paragraph{Dataset}
\label{exp:dataset}
To align with prior work~\citep{luo20213d, peng2022pocket2mol}, we use CrossDocked2020 ~\citep{francoeur2020three} to fine-tune and evaluate our model. We adopt the preprocessing and splitting procedure described in ~\citep{luo20213d}. Starting from 22.5M docked protein–ligand complexes, we keep only those with RMSD to the ground truth below 1\AA\ and with protein sequence identity under 30\%. This results in a curated set of 100{,}000 complexes for training and 100 proteins reserved for testing. The training set is further divided for SFT and DPO (see Section~\ref{sec:training_strategy}).

\paragraph{Baselines}
We benchmark \ours{} against a range of representative baselines for target-aware molecular generation: prior structure-based models (\textbf{liGAN}~\citep{ragoza2022generating}, \textbf{GraphBP}~\citep{liu2022generating}); autoregressive approaches (\textbf{AR}~\citep{luo20213d}, \textbf{Pocket2Mol}~\citep{peng2022pocket2mol}, \textbf{TamGen}~\citep{wu2024tamgen}); diffusion-based methods (\textbf{TargetDiff}~\citep{guan20233d}, \textbf{DecompDiff}~\citep{guan2023decompdiff}); the BFN-based \textbf{MolCRAFT}~\citep{qu2024molcraft}; and the flow-based \textbf{FlowSBDD}~\citep{zhang2024rectified}. Together, these baselines span diverse methodological families and provide a balanced foundation for evaluating the effectiveness of \ours{}.

\paragraph{Evaluation} To provide a comprehensive assessment of generated molecules in drug design applications, we consider the following evaluation metrics: (1) \textbf{Vina Dock}, denoting the binding affinity score estimated via re-docking; (2) \textbf{High Affinity}, measuring for each pocket the fraction of generated molecules that achieve Vina Dock scores no worse than the corresponding test-set ligands;(3) \textbf{QED} (Quantitative Estimate of Drug-likeness)~\citep{bickerton2012quantifying} ;(4) \textbf{SA} (Synthetic Accessibility)~\citep{ertl2009estimation, you2018graph} ;(5) \textbf{Diversity}, computed as the average pairwise Tanimoto similarity among generated molecules within each pocket;(6) \textbf{Success Rate}, representing the fraction of molecules that are drug-like, synthesizable, and high-affinity binders, is computed following~\citep{long2022zero} and~\citep{guan2023decompdiff} as the proportion of molecules with QED $>$ 0.25, SA $>$ 0.59, and Vina Dock $< -8.18$. To evaluate binding affinity to the target, we use AutoDock Vina~\citep{eberhardt2021autodock}, adopting the evaluation protocol described by~\citep{guan20233d}. For each protein pocket, we evaluate 100 generated molecules.

\subsection{Main Results}

\begin{table}[!t]
\caption{Summary of molecular properties between \ours{} and other baseline methods for pocket-aware drug design. ($\uparrow$) / ($\downarrow$) indicates larger / smaller is better. Top-2 results are marked in \textbf{bold} and \underline{underline}, respectively.}
\label{crossdocked2020-main-results}
\centering
\resizebox{\linewidth}{!}{
\begin{tabular}{lcccccccc}
\toprule
\multicolumn{1}{l}{Methods} & \multicolumn{1}{c}{Vina Dock ($\downarrow$)} & \multicolumn{1}{c}{High Affinity ($\uparrow$)} & \multicolumn{1}{c}{QED ($\uparrow$)} & \multicolumn{1}{c}{SA ($\uparrow$)} & \multicolumn{1}{c}{Diversity ($\uparrow$)} & \multicolumn{1}{c}{Success Rate ($\uparrow$)} \\ 
\midrule
Reference & -7.45 & - & 0.48 & 0.73 & - & 25.0\% \\ 
\midrule
LiGAN & -6.33 & 21.1\% & 0.39 & 0.59 & 0.66 & 3.9\% \\
GraphBP & -4.80 & 14.2\% & 0.43 & 0.49 & \textbf{0.79} & 0.1\% \\
AR & -6.75 & 37.9\% & \underline{0.51} & 0.63 & 0.70 & 7.1\% \\ 
Pocket2Mol & -7.15 & 48.4\% & \textbf{0.56} & 0.74 & 0.69 & 24.4\% \\ 
TamGen & -7.48 & 52.6\% & \textbf{0.56} & \underline{0.77} & 0.75 & 32.4\% \\
TargetDiff & -7.80 & 58.1\% & 0.48 & 0.58 & 0.72 & 10.5\% \\ 
DecompDiff & -8.39 & 64.4\% & 0.45 & 0.61 & 0.68 & 24.5\% \\
MolCRAFT & -7.92 & 59.1\% & 0.50 & 0.69 & 0.72 & 26.8\% \\
FlowSBDD & -8.50 & 63.4\% & 0.47 & 0.51 & 0.75 & - \\
\midrule
\ours{} & -7.62 & 55.1\% & \textbf{0.56} & \underline{0.77} & \underline{0.76} & 33.2\% \\ 
\ours{}-RL$^{\rm dock}$ & \textbf{-9.29} & \textbf{83.7\%} & 0.44 & \underline{0.77} & 0.63 & \textbf{59.3\%} \\
\ours{}-RL & \underline{-8.59} & \underline{74.6\%} & \textbf{0.56} & \textbf{0.78} & 0.71 & \underline{53.4\%} \\ 
\bottomrule
\end{tabular}
}
\end{table}

Table~\ref{crossdocked2020-main-results} summarizes the performance of \ours{} and its RL variants, including \ours{}-RL and \ours{}-RL$^{\rm dock}$, where the latter denotes the model optimized with DPO solely for affinity. Overall, \ours{} outperforms all baselines in five key metrics: Vina Dock and High Affinity for binding affinity, QED, SA, and Success Rate for molecular properties, while also maintaining competitive diversity. For binding affinity, the RL-enhanced model achieves the best Vina Dock score and the highest High Affinity, being the first to surpass the 70\% threshold and outperforming strong baselines such as FlowSBDD and DecompDiff. Moreover, our gains are substantially larger than those of autoregressive methods, underscoring the importance of the structure encoder in capturing and incorporating structural information. 

For molecular properties, both \ours{} and \ours{}-RL establish state-of-the-art results. On QED, our models perform comparably with strong autoregressive baselines such as Pocket2Mol~\citep{peng2022pocket2mol} and TamGen~\citep{wu2024tamgen}, while achieving the highest SA score (0.78), clearly outperforming diffusion- and flow-based methods. Most importantly, \ours{}-RL attains a high Success Rate, reflecting its ability to jointly optimize binding affinity and drug-likeness. These results highlight that our approach effectively leverages the strengths of autoregressive modeling while extending them to drug-like and synthesizable molecule generation. For diversity, \ours{} achieves 0.76, second only to the early method GraphBP, which performs poorly on affinity and molecular properties. With RL, diversity decreases slightly—a trade-off also observed in prior works~\citep{cheng2024decomposed}—but remains above 0.70, indicating that our RL improves affinity while still preserving meaningful variation in generation.

\begin{figure}[!ht]
    \centering
    \includegraphics[width=0.95\linewidth]{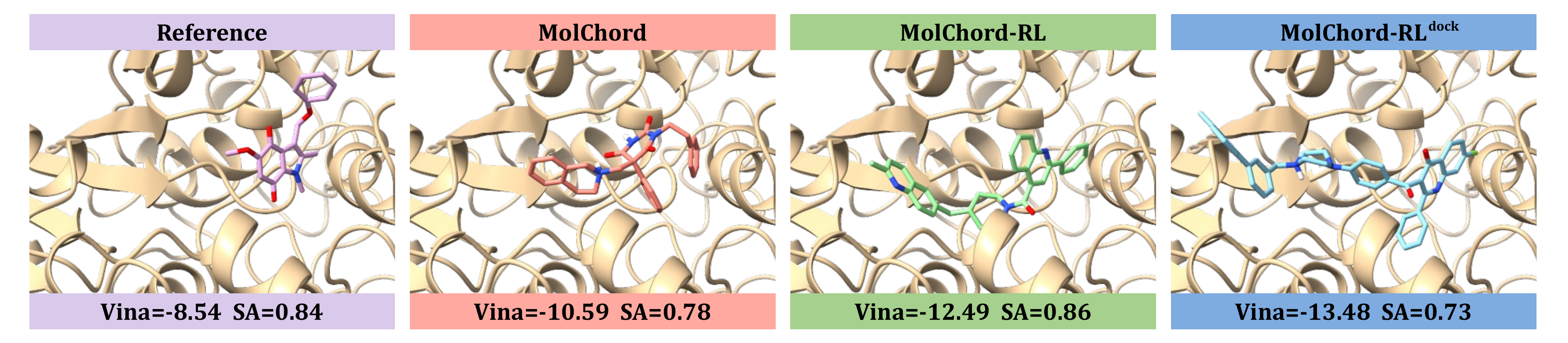}
    \caption{Visualizations of reference molecules and ligands generated by \ours{}, \ours{}-RL, and \ours{}-RL$^{\rm dock}$ for protein pocket 1gg5. Vina score and SA are reported.}
    \label{fig:case-study}
\end{figure}

Notably, the performance of \ours{}-RL$^{\rm dock}$ (Table~\ref{crossdocked2020-main-results}) highlights the trade-off of DPO. While DPO is capable of aggressively improving Vina Dock scores and maintain state-of-the-art SA, it incurs acceptable declines in QED and diversity. Our design instead prioritizes balance, leveraging reward shaping to jointly enforce binding affinity, pharmacological properties, and molecular diversity, achieving strong and stable performance across objectives. Figure~\ref{fig:case-study} provides case studies comparing reference molecules with ligands generated by our approach. We observe that (i) \ours{} produces candidates with strong overall quality, (ii) \ours{}-RL simultaneously improves binding affinity and molecular properties, and (iii) \ours{}-RL$^{\rm dock}$ achieves high affinity but at the expense of molecular attributes. These examples further illustrate the advantage of balanced optimization in our approach.

\begin{figure}[t]
    \centering
    \begin{minipage}[t]{0.49\linewidth}
        \centering
        \includegraphics[width=\linewidth]{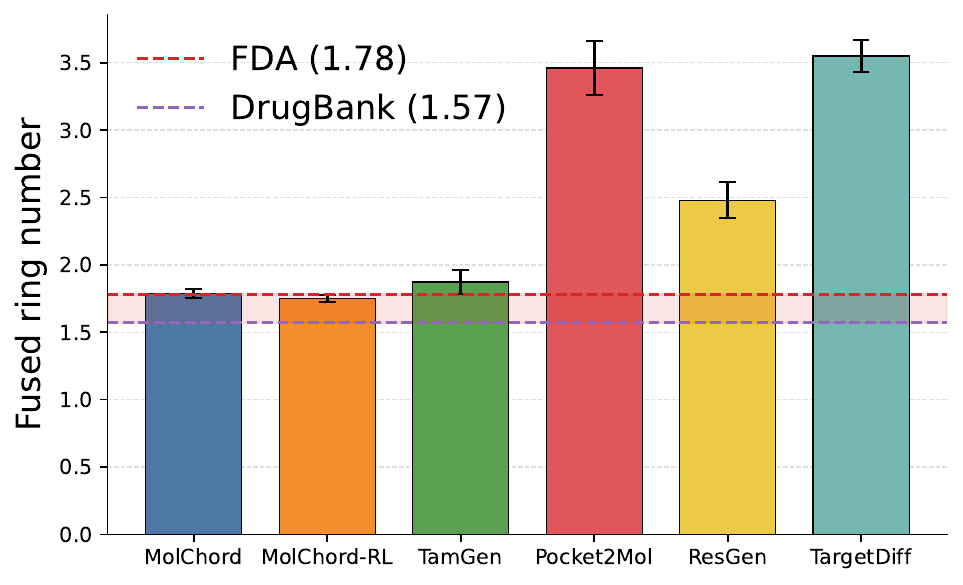}
        \caption{Barplot of the number of fused rings in top-ranked compounds generated by representative methods. For each method, statistics of 1,000 compounds (100 targets × 10 compounds with the highest docking scores) are reported.}
        \label{fused-rings}
    \end{minipage}
    \hfill
    \begin{minipage}[t]{0.49\linewidth}
        \centering
        \includegraphics[width=\linewidth]{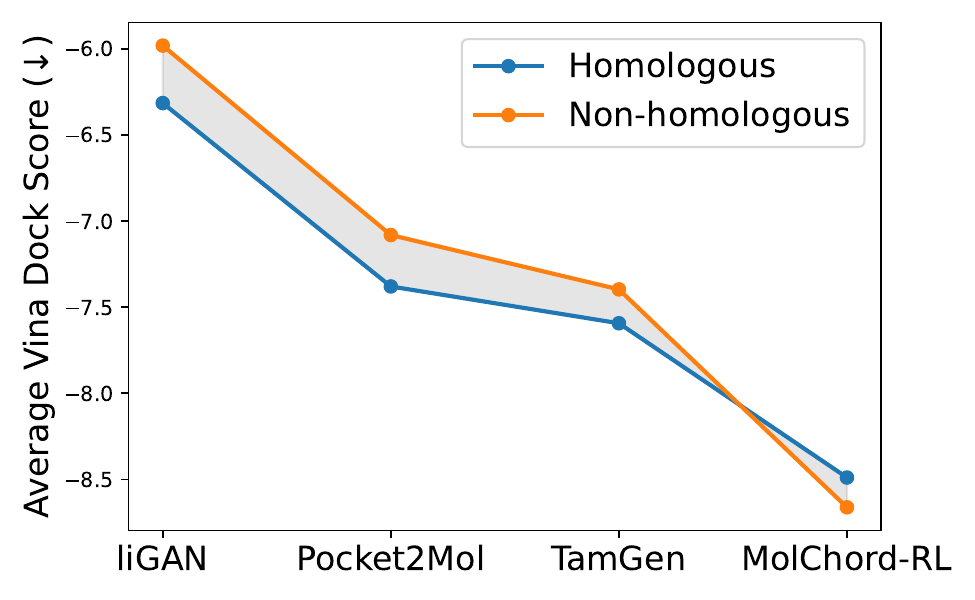}
        \caption{OOD generalization: average Vina Dock scores on homologous vs non-homologous proteins for representative methods.}
        \label{ood-results}
    \end{minipage}
\end{figure}

\paragraph{Fused Ring} 
\label{4:fused-rings}
Fused rings refer to ring systems in which two or more rings share atoms, a structural motif commonly found in bioactive molecules, and often influence both binding and drug-likeness. While fused rings can contribute to favorable binding poses, an excessive number of fused rings is undesirable: prior work such as TamGen~\citep{wu2024tamgen} has shown that an excessive number of fused rings may lead to lower synthetic accessibility~\citep{skoraczynski2023critical, ertl2009estimation, peng2023moldiff}, increased cellular toxicity, and decreased developability~\citep{peng2023moldiff, ritchie2009impact}. Indeed, fused rings are known to correlate with QED and SA, making them a useful proxy for chemical plausibility. 

Figure~\ref{fused-rings} shows that our method is the first to match the range of fused ring of approved drugs (see Appendix Table~\ref{tab:fused-rings} for detailed statistics): \ours{} achieves an average of 1.79, close to the FDA reference (1.78), and \ours{}-RL further improves to 1.75. For context, DrugBank averages 1.57 fused rings, while representative baselines such as Pocket2Mol, TargetDiff, and ResGen substantially overproduce complex ring systems. These results demonstrate that our approach generates not only high-affinity molecules but also chemically plausible and pharmaceutically relevant candidates, with RL fine-tuning providing additional regularization.

\paragraph{Out-of-distribution generalization} 

To further evaluate generalization, we split test proteins into \emph{homologous} and \emph{non-homologous} subsets based on sequence identity with the training set. Pairwise identities were computed using MMseqs2~\citep{steinegger2017mmseqs2}, and proteins sharing more than 30\% identity with any training sequence were classified as homologous, yielding 40 homologous and 60 non-homologous cases. Figure~\ref{ood-results} reports average Vina Dock scores for both subsets (see Appendix Table~\ref{tab:homology-ablation} for detailed statistics). Prior methods such as liGAN~\citep{ragoza2022generating}, Pocket2Mol~\citep{peng2022pocket2mol}, and TamGen~\citep{wu2024tamgen} show clear performance drops on non-homologous proteins. In contrast, \ours{}-RL not only maintains performance but improves when generalizing to non-homologous proteins ($-8.49 \rightarrow -8.66$, improvement of $+0.17$). We attribute this robustness to the structure encoder, which leverages large-scale pretraining to capture transferable structural features. These results highlight that our approach generalizes beyond training distributions, a critical requirement for real-world drug discovery.

\subsection{Ablation Study}

\paragraph{Effect of Structure-Sequence Alignment}  
We further examine the role of alignment design by comparing three variants: 
(i) \textbf{Naïve Alignment}, which directly uses the CrossDocked2020~\citep{francoeur2020three} training set for alignment; 
(ii) \textbf{Protein–FASTA Alignment}, which performs alignment solely through protein structure–to–FASTA mapping; and 
(iii) \textbf{Full Alignment}, our complete model with molecule–protein–complex alignment.
These variants differ only in Stage~A, while Stage~B training is kept identical across settings, and all models are evaluated on the CrossDocked2020 benchmark for comparison.
As shown in Table~\ref{crossdocked2020-ablation-1}, the full alignment achieves the strongest overall performance. For (i) Naïve Alignment, using downstream training dataset directly for Stage~A and Stage~B leads to overfitting: proteins are not well aligned across structure and sequence space, and the limited chemical exploration results in weaker docking scores and reduced diversity. For (ii) Protein–FASTA Alignment, which aligns proteins at the structure–sequence level and thus alleviates overfitting by better capturing structural–sequential consistency. However, the absence of molecule-related and protein-to-annotation tasks limits chemical space exploration and reduces the benefit of leveraging textual alignment signals. In contrast, (iii) Full Alignment combines protein, molecule, and complex supervision, resulting in the strongest binding affinity and molecular properties. These results highlight the importance of a comprehensive alignment strategy that integrates multiple sources of supervision.

\begin{table}[!ht]
\caption{The influence of Structure-Sequence Alignment}
\label{crossdocked2020-ablation-1}
\centering
\resizebox{\linewidth}{!}{
\begin{tabular}{lccccccccc}
\toprule
\multicolumn{1}{l}{Setting} & \multicolumn{1}{c}{Vina Dock ($\downarrow$)} & \multicolumn{1}{c}{High Affinity ($\uparrow$)} & \multicolumn{1}{c}{QED ($\uparrow$)} & \multicolumn{1}{c}{SA ($\uparrow$)} & \multicolumn{1}{c}{Diversity ($\uparrow$)} & \multicolumn{1}{c}{Success Rate ($\uparrow$)} \\ 
\midrule
Naïve & -7.38 & 49.8\% & 0.55 & 0.77 & 0.74 & 28.6\% \\
Protein–FASTA & -7.44 & 50.7\% & 0.57 & 0.77 & 0.74 & 31.2\% \\
Full & -7.62 & 54.7\% & 0.56 & 0.77 & 0.76 & 33.2\% \\
\bottomrule
\end{tabular}
}
\end{table}

\paragraph{Effect of data partitioning}
We conduct ablations to disentangle the effect of stratified data usage in SFT and DPO, with results summarized in Table~\ref{crossdocked2020-ablation-2}. First, let $\mathcal{D}_{\text{full}}$ denote the entire curated CrossDocked2020 dataset. Comparing SFT trained on $\mathcal{D}_{\text{full}}$ versus on the stratified subset $\mathcal{D}_{\text{B}}$, we observe only minor differences: a slight decrease in affinity, accompanied by modest gains in SA and diversity. This indicates that the partitioning procedure itself has limited impact on supervised learning.
Second, we investigate the effect of partitioning on preference optimization. Recall that after diversity-based filtering, the dataset used for DPO is denoted as $\mathcal{D}_{\text{DPO}}$.
We compare three settings: 
(i) SFT($\mathcal{D}_{\text{full}}$)+DPO(random), 
(ii) SFT($\mathcal{D}_{\text{B}}$)+DPO(random), and 
(iii) SFT($\mathcal{D}_{\text{B}}$)+DPO($\mathcal{D}_{\text{DPO}}$), 
where “random’’ denotes a subset drawn uniformly at random from $\mathcal{D}^{\text{pool}}_{\text{DPO}}$ with the same size as $\mathcal{D}_{\text{DPO}}$. 
This comparison reveals two effects. First, comparing (i) and (ii), we find that separating the preference pool from SFT data yields better DPO performance, with clear gains in affinity. Second, comparing (ii) and (iii), our diversity-based filtering strategy proves effective, resulting in consistent improvements across affinity, molecular properties, and diversity.

\begin{table}[!ht]
\caption{The influence of data partitioning}
\label{crossdocked2020-ablation-2}
\centering
\resizebox{\linewidth}{!}{
\begin{tabular}{lccccccccc}
\toprule
\multicolumn{1}{l}{Setting} & \multicolumn{1}{c}{Vina Dock ($\downarrow$)} & \multicolumn{1}{c}{High Affinity ($\uparrow$)} & \multicolumn{1}{c}{QED ($\uparrow$)} & \multicolumn{1}{c}{SA ($\uparrow$)} & \multicolumn{1}{c}{Diversity ($\uparrow$)} & \multicolumn{1}{c}{Success Rate ($\uparrow$)} \\ 
\midrule
SFT($\mathcal{D}_{\text{full}}$) & -7.64 & 55.1\% & 0.56 & 0.78 & 0.75 & 33.5\% \\
SFT(\(\mathcal{D}_{\text{B}}\)) & -7.62 & 54.7\% & 0.56 & 0.77 & 0.76 & 33.2\% \\
\midrule
SFT($\mathcal{D}_{\text{full}}$)+DPO(random) & -8.22 & 67.5\% & 0.54 & 0.77 & 0.68 & 42.1\% \\
SFT(\(\mathcal{D}_{\text{B}}\))+DPO(random) & -8.44 & 71.6\% & 0.53 & 0.77 & 0.68 & 47.1\% \\
SFT(\(\mathcal{D}_{\text{B}}\))+DPO(\(\mathcal{D}_{\text{DPO}}\)) & -8.59 & 74.6\% & 0.56 & 0.78 & 0.71 & 53.4\% \\
\bottomrule
\end{tabular}
}
\end{table}

\section{Conclusion}

In this paper, we introduced \ours{}, a framework for SBDD that combines a diffusion-based structure encoder with an autoregressive generator. The framework enhances alignment by linking proteins with FASTA and descriptions, molecules with SMILES and descriptions, and complexes with paired FASTA–SMILES representations. To enable effective preference optimization, we proposed a stratified data split and constructed a curated DPO dataset, which proved critical for improving model performance. 
Beyond binding affinity, our method effectively balances diversity and pharmacological properties, both of which are crucial for drug discovery. The results highlight the potential of our approach as a general framework for SBDD.

\bibliographystyle{iclr2026_conference}
\bibliography{reference}

\begin{thebibliography}{55}
\providecommand{\natexlab}[1]{#1}
\providecommand{\url}[1]{\texttt{#1}}
\expandafter\ifx\csname urlstyle\endcsname\relax
  \providecommand{\doi}[1]{doi: #1}\else
  \providecommand{\doi}{doi: \begingroup \urlstyle{rm}\Url}\fi

\bibitem[Abramson et~al.(2024)Abramson, Adler, Dunger, Evans, Green, Pritzel, Ronneberger, Willmore, Ballard, Bambrick, et~al.]{abramson2024accurate}
Josh Abramson, Jonas Adler, Jack Dunger, Richard Evans, Tim Green, Alexander Pritzel, Olaf Ronneberger, Lindsay Willmore, Andrew~J Ballard, Joshua Bambrick, et~al.
\newblock Accurate structure prediction of biomolecular interactions with alphafold 3.
\newblock \emph{Nature}, 630\penalty0 (8016):\penalty0 493--500, 2024.

\bibitem[Anderson(2003)]{anderson2003process}
Amy~C Anderson.
\newblock The process of structure-based drug design.
\newblock \emph{Chemistry \& biology}, 10\penalty0 (9):\penalty0 787--797, 2003.

\bibitem[Batool et~al.(2019)Batool, Ahmad, and Choi]{batool2019structure}
Maria Batool, Bilal Ahmad, and Sangdun Choi.
\newblock A structure-based drug discovery paradigm.
\newblock \emph{International journal of molecular sciences}, 20\penalty0 (11):\penalty0 2783, 2019.

\bibitem[Berman et~al.(2000)Berman, Westbrook, Feng, Gilliland, Bhat, Weissig, Shindyalov, and Bourne]{berman2000protein}
Helen~M Berman, John Westbrook, Zukang Feng, Gary Gilliland, Talapady~N Bhat, Helge Weissig, Ilya~N Shindyalov, and Philip~E Bourne.
\newblock The protein data bank.
\newblock \emph{Nucleic acids research}, 28\penalty0 (1):\penalty0 235--242, 2000.

\bibitem[Bickerton et~al.(2012)Bickerton, Paolini, Besnard, Muresan, and Hopkins]{bickerton2012quantifying}
G~Richard Bickerton, Gaia~V Paolini, J{\'e}r{\'e}my Besnard, Sorel Muresan, and Andrew~L Hopkins.
\newblock Quantifying the chemical beauty of drugs.
\newblock \emph{Nature chemistry}, 4\penalty0 (2):\penalty0 90--98, 2012.

\bibitem[Boutet et~al.(2007)Boutet, Lieberherr, Tognolli, Schneider, and Bairoch]{boutet2007uniprotkb}
Emmanuel Boutet, Damien Lieberherr, Michael Tognolli, Michel Schneider, and Amos Bairoch.
\newblock Uniprotkb/swiss-prot: the manually annotated section of the uniprot knowledgebase.
\newblock In \emph{Plant bioinformatics: methods and protocols}, pp.\  89--112. Springer, 2007.

\bibitem[Cheng et~al.(2024)Cheng, Zhou, Yang, Bao, and Gu]{cheng2024decomposed}
Xiwei Cheng, Xiangxin Zhou, Yuwei Yang, Yu~Bao, and Quanquan Gu.
\newblock Decomposed direct preference optimization for structure-based drug design.
\newblock \emph{arXiv preprint arXiv:2407.13981}, 2024.

\bibitem[Davies et~al.(2006)Davies, Glick, and Jenkins]{davies2006streamlining}
John~W Davies, Meir Glick, and Jeremy~L Jenkins.
\newblock Streamlining lead discovery by aligning in silico and high-throughput screening.
\newblock \emph{Current opinion in chemical biology}, 10\penalty0 (4):\penalty0 343--351, 2006.

\bibitem[DiMasi et~al.(2016)DiMasi, Grabowski, and Hansen]{dimasi2016innovation}
Joseph~A DiMasi, Henry~G Grabowski, and Ronald~W Hansen.
\newblock Innovation in the pharmaceutical industry: new estimates of r\&d costs.
\newblock \emph{Journal of health economics}, 47:\penalty0 20--33, 2016.

\bibitem[Dubey et~al.(2024)Dubey, Jauhri, Pandey, Kadian, Al-Dahle, Letman, Mathur, Schelten, Yang, Fan, et~al.]{dubey2024llama}
Abhimanyu Dubey, Abhinav Jauhri, Abhinav Pandey, Abhishek Kadian, Ahmad Al-Dahle, Aiesha Letman, Akhil Mathur, Alan Schelten, Amy Yang, Angela Fan, et~al.
\newblock The llama 3 herd of models.
\newblock \emph{arXiv e-prints}, pp.\  arXiv--2407, 2024.

\bibitem[Eberhardt et~al.(2021)Eberhardt, Santos-Martins, Tillack, and Forli]{eberhardt2021autodock}
Jerome Eberhardt, Diogo Santos-Martins, Andreas~F Tillack, and Stefano Forli.
\newblock Autodock vina 1.2. 0: new docking methods, expanded force field, and python bindings.
\newblock \emph{Journal of chemical information and modeling}, 61\penalty0 (8):\penalty0 3891--3898, 2021.

\bibitem[Ertl \& Schuffenhauer(2009)Ertl and Schuffenhauer]{ertl2009estimation}
Peter Ertl and Ansgar Schuffenhauer.
\newblock Estimation of synthetic accessibility score of drug-like molecules based on molecular complexity and fragment contributions.
\newblock \emph{Journal of cheminformatics}, 1\penalty0 (1):\penalty0 8, 2009.

\bibitem[Feng et~al.(2023)Feng, Li, Jia, Ma, and Lan]{feng2023protein}
Shikun Feng, Minghao Li, Yinjun Jia, Weiying Ma, and Yanyan Lan.
\newblock Protein-ligand binding representation learning from fine-grained interactions.
\newblock \emph{arXiv preprint arXiv:2311.16160}, 2023.

\bibitem[Feng et~al.(2024)Feng, Wang, Lin, Zhu, Wang, Dong, Bai, Wang, Zhou, Peng, et~al.]{feng2024generation}
Wei Feng, Lvwei Wang, Zaiyun Lin, Yanhao Zhu, Han Wang, Jianqiang Dong, Rong Bai, Huting Wang, Jielong Zhou, Wei Peng, et~al.
\newblock Generation of 3d molecules in pockets via a language model.
\newblock \emph{Nature Machine Intelligence}, 6\penalty0 (1):\penalty0 62--73, 2024.

\bibitem[Flam-Shepherd \& Aspuru-Guzik(2023)Flam-Shepherd and Aspuru-Guzik]{flam2023language}
Daniel Flam-Shepherd and Al{\'a}n Aspuru-Guzik.
\newblock Language models can generate molecules, materials, and protein binding sites directly in three dimensions as xyz, cif, and pdb files.
\newblock \emph{arXiv preprint arXiv:2305.05708}, 2023.

\bibitem[Francoeur et~al.(2020)Francoeur, Masuda, Sunseri, Jia, Iovanisci, Snyder, and Koes]{francoeur2020three}
Paul~G Francoeur, Tomohide Masuda, Jocelyn Sunseri, Andrew Jia, Richard~B Iovanisci, Ian Snyder, and David~R Koes.
\newblock Three-dimensional convolutional neural networks and a cross-docked data set for structure-based drug design.
\newblock \emph{Journal of chemical information and modeling}, 60\penalty0 (9):\penalty0 4200--4215, 2020.

\bibitem[Gao et~al.(2023)Gao, Jia, Mo, Ni, Ma, Ma, and Lan]{gao2023profsa}
Bowen Gao, Yinjun Jia, Yuanle Mo, Yuyan Ni, Weiying Ma, Zhiming Ma, and Yanyan Lan.
\newblock Profsa: Self-supervised pocket pretraining via protein fragment-surroundings alignment.
\newblock \emph{arXiv preprint arXiv:2310.07229}, 2023.

\bibitem[Gu et~al.(2024)Gu, Xu, Powers, Nie, Geffner, Kreis, Leskovec, Vahdat, and Ermon]{gu2024aligning}
Siyi Gu, Minkai Xu, Alexander Powers, Weili Nie, Tomas Geffner, Karsten Kreis, Jure Leskovec, Arash Vahdat, and Stefano Ermon.
\newblock Aligning target-aware molecule diffusion models with exact energy optimization.
\newblock \emph{Advances in Neural Information Processing Systems}, 37:\penalty0 44040--44063, 2024.

\bibitem[Guan et~al.(2023{\natexlab{a}})Guan, Qian, Peng, Su, Peng, and Ma]{guan20233d}
Jiaqi Guan, Wesley~Wei Qian, Xingang Peng, Yufeng Su, Jian Peng, and Jianzhu Ma.
\newblock 3d equivariant diffusion for target-aware molecule generation and affinity prediction.
\newblock \emph{arXiv preprint arXiv:2303.03543}, 2023{\natexlab{a}}.

\bibitem[Guan et~al.(2023{\natexlab{b}})Guan, Zhou, Yang, Bao, Peng, Ma, Liu, Wang, and Gu]{guan2023decompdiff}
Jiaqi Guan, Xiangxin Zhou, Yuwei Yang, Yu~Bao, Jian Peng, Jianzhu Ma, Qiang Liu, Liang Wang, and Quanquan Gu.
\newblock Decompdiff: Diffusion models with decomposed priors for structure-based drug design.
\newblock In \emph{International Conference on Machine Learning}, pp.\  11827--11846. PMLR, 2023{\natexlab{b}}.

\bibitem[He et~al.(2025)He, Fang, Shan, Pan, Wei, Chen, Chen, Liu, Zeng, Zhou, et~al.]{he2025generalized}
Yong He, Pan Fang, Yongtao Shan, Yuanfei Pan, Yanhong Wei, Yichang Chen, Yihao Chen, Yi~Liu, Zhenyu Zeng, Zhan Zhou, et~al.
\newblock Generalized biological foundation model with unified nucleic acid and protein language.
\newblock \emph{Nature Machine Intelligence}, pp.\  1--12, 2025.

\bibitem[Hu et~al.(2025)Hu, Liu, Chen, Zhao, Zhang, and Liu]{hu20253dmolformer}
Xiuyuan Hu, Guoqing Liu, Can Chen, Yang Zhao, Hao Zhang, and Xue Liu.
\newblock 3dmolformer: A dual-channel framework for structure-based drug discovery.
\newblock \emph{arXiv preprint arXiv:2502.05107}, 2025.

\bibitem[Huang \& Zhang(2025)Huang and Zhang]{huang2025molform}
Jie Huang and Daiheng Zhang.
\newblock Molform: Multi-modal flow matching for structure-based drug design.
\newblock \emph{arXiv preprint arXiv:2507.05503}, 2025.

\bibitem[Karras et~al.(2022)Karras, Aittala, Aila, and Laine]{karras2022elucidating}
Tero Karras, Miika Aittala, Timo Aila, and Samuli Laine.
\newblock Elucidating the design space of diffusion-based generative models.
\newblock \emph{Advances in neural information processing systems}, 35:\penalty0 26565--26577, 2022.

\bibitem[Liu et~al.(2022)Liu, Luo, Uchino, Maruhashi, and Ji]{liu2022generating}
Meng Liu, Youzhi Luo, Kanji Uchino, Koji Maruhashi, and Shuiwang Ji.
\newblock Generating 3d molecules for target protein binding.
\newblock \emph{arXiv preprint arXiv:2204.09410}, 2022.

\bibitem[Liu et~al.(2023)Liu, Zhang, Xia, Wu, Xie, Qin, Zhang, and Liu]{liu2023molxpt}
Zequn Liu, Wei Zhang, Yingce Xia, Lijun Wu, Shufang Xie, Tao Qin, Ming Zhang, and Tie-Yan Liu.
\newblock Molxpt: Wrapping molecules with text for generative pre-training.
\newblock In \emph{The 61st Annual Meeting Of The Association For Computational Linguistics}, 2023.

\bibitem[Long et~al.(2022)Long, Zhou, Dai, and Zhou]{long2022zero}
Siyu Long, Yi~Zhou, Xinyu Dai, and Hao Zhou.
\newblock Zero-shot 3d drug design by sketching and generating.
\newblock \emph{Advances in Neural Information Processing Systems}, 35:\penalty0 23894--23907, 2022.

\bibitem[Luo et~al.(2021)Luo, Guan, Ma, and Peng]{luo20213d}
Shitong Luo, Jiaqi Guan, Jianzhu Ma, and Jian Peng.
\newblock A 3d generative model for structure-based drug design.
\newblock \emph{Advances in Neural Information Processing Systems}, 34:\penalty0 6229--6239, 2021.

\bibitem[Nakata et~al.(2023)Nakata, Mori, and Tanaka]{nakata2023end}
Shuya Nakata, Yoshiharu Mori, and Shigenori Tanaka.
\newblock End-to-end protein--ligand complex structure generation with diffusion-based generative models.
\newblock \emph{BMC bioinformatics}, 24\penalty0 (1):\penalty0 233, 2023.

\bibitem[O'Boyle et~al.(2011)O'Boyle, Banck, James, Morley, Vandermeersch, and Hutchison]{o2011open}
Noel~M O'Boyle, Michael Banck, Craig~A James, Chris Morley, Tim Vandermeersch, and Geoffrey~R Hutchison.
\newblock Open babel: An open chemical toolbox.
\newblock \emph{Journal of cheminformatics}, 3\penalty0 (1):\penalty0 33, 2011.

\bibitem[Ouyang et~al.(2022)Ouyang, Wu, Jiang, Almeida, Wainwright, Mishkin, Zhang, Agarwal, Slama, Ray, et~al.]{ouyang2022training}
Long Ouyang, Jeffrey Wu, Xu~Jiang, Diogo Almeida, Carroll Wainwright, Pamela Mishkin, Chong Zhang, Sandhini Agarwal, Katarina Slama, Alex Ray, et~al.
\newblock Training language models to follow instructions with human feedback.
\newblock \emph{Advances in neural information processing systems}, 35:\penalty0 27730--27744, 2022.

\bibitem[Paul et~al.(2010)Paul, Mytelka, Dunwiddie, Persinger, Munos, Lindborg, and Schacht]{paul2010improve}
Steven~M Paul, Daniel~S Mytelka, Christopher~T Dunwiddie, Charles~C Persinger, Bernard~H Munos, Stacy~R Lindborg, and Aaron~L Schacht.
\newblock How to improve r\&d productivity: the pharmaceutical industry's grand challenge.
\newblock \emph{Nature reviews Drug discovery}, 9\penalty0 (3):\penalty0 203--214, 2010.

\bibitem[Peebles \& Xie(2023)Peebles and Xie]{peebles2023scalable}
William Peebles and Saining Xie.
\newblock Scalable diffusion models with transformers.
\newblock In \emph{Proceedings of the IEEE/CVF international conference on computer vision}, pp.\  4195--4205, 2023.

\bibitem[Peng et~al.(2022)Peng, Luo, Guan, Xie, Peng, and Ma]{peng2022pocket2mol}
Xingang Peng, Shitong Luo, Jiaqi Guan, Qi~Xie, Jian Peng, and Jianzhu Ma.
\newblock Pocket2mol: Efficient molecular sampling based on 3d protein pockets.
\newblock In \emph{International conference on machine learning}, pp.\  17644--17655. PMLR, 2022.

\bibitem[Peng et~al.(2023)Peng, Guan, Liu, and Ma]{peng2023moldiff}
Xingang Peng, Jiaqi Guan, Qiang Liu, and Jianzhu Ma.
\newblock Moldiff: Addressing the atom-bond inconsistency problem in 3d molecule diffusion generation.
\newblock \emph{arXiv preprint arXiv:2305.07508}, 2023.

\bibitem[Polishchuk et~al.(2013)Polishchuk, Madzhidov, and Varnek]{polishchuk2013estimation}
Pavel~G Polishchuk, Timur~I Madzhidov, and Alexandre Varnek.
\newblock Estimation of the size of drug-like chemical space based on gdb-17 data.
\newblock \emph{Journal of computer-aided molecular design}, 27\penalty0 (8):\penalty0 675--679, 2013.

\bibitem[Qu et~al.(2024)Qu, Qiu, Song, Gong, Han, Zheng, Zhou, and Ma]{qu2024molcraft}
Yanru Qu, Keyue Qiu, Yuxuan Song, Jingjing Gong, Jiawei Han, Mingyue Zheng, Hao Zhou, and Wei-Ying Ma.
\newblock Molcraft: structure-based drug design in continuous parameter space.
\newblock \emph{arXiv preprint arXiv:2404.12141}, 2024.

\bibitem[Rafailov et~al.(2023)Rafailov, Sharma, Mitchell, Manning, Ermon, and Finn]{rafailov2023direct}
Rafael Rafailov, Archit Sharma, Eric Mitchell, Christopher~D Manning, Stefano Ermon, and Chelsea Finn.
\newblock Direct preference optimization: Your language model is secretly a reward model.
\newblock \emph{Advances in neural information processing systems}, 36:\penalty0 53728--53741, 2023.

\bibitem[Ragoza et~al.(2022)Ragoza, Masuda, and Koes]{ragoza2022generating}
Matthew Ragoza, Tomohide Masuda, and David~Ryan Koes.
\newblock Generating 3d molecules conditional on receptor binding sites with deep generative models.
\newblock \emph{Chemical science}, 13\penalty0 (9):\penalty0 2701--2713, 2022.

\bibitem[Ritchie \& Macdonald(2009)Ritchie and Macdonald]{ritchie2009impact}
Timothy~J Ritchie and Simon~JF Macdonald.
\newblock The impact of aromatic ring count on compound developability--are too many aromatic rings a liability in drug design?
\newblock \emph{Drug discovery today}, 14\penalty0 (21-22):\penalty0 1011--1020, 2009.

\bibitem[Schneider et~al.(2020)Schneider, Walters, Plowright, Sieroka, Listgarten, Goodnow~Jr, Fisher, Jansen, Duca, Rush, et~al.]{schneider2020rethinking}
Petra Schneider, W~Patrick Walters, Alleyn~T Plowright, Norman Sieroka, Jennifer Listgarten, Robert~A Goodnow~Jr, Jasmin Fisher, Johanna~M Jansen, Jos{\'e}~S Duca, Thomas~S Rush, et~al.
\newblock Rethinking drug design in the artificial intelligence era.
\newblock \emph{Nature reviews drug discovery}, 19\penalty0 (5):\penalty0 353--364, 2020.

\bibitem[Schneuing et~al.(2024)Schneuing, Harris, Du, Didi, Jamasb, Igashov, Du, Gomes, Blundell, Lio, et~al.]{schneuing2024structure}
Arne Schneuing, Charles Harris, Yuanqi Du, Kieran Didi, Arian Jamasb, Ilia Igashov, Weitao Du, Carla Gomes, Tom~L Blundell, Pietro Lio, et~al.
\newblock Structure-based drug design with equivariant diffusion models.
\newblock \emph{Nature Computational Science}, 4\penalty0 (12):\penalty0 899--909, 2024.

\bibitem[Skoraczy{\'n}ski et~al.(2023)Skoraczy{\'n}ski, Kitlas, Miasojedow, and Gambin]{skoraczynski2023critical}
Grzegorz Skoraczy{\'n}ski, Mateusz Kitlas, B{\l}a{\.z}ej Miasojedow, and Anna Gambin.
\newblock Critical assessment of synthetic accessibility scores in computer-assisted synthesis planning.
\newblock \emph{Journal of Cheminformatics}, 15\penalty0 (1):\penalty0 6, 2023.

\bibitem[Steinegger \& S{\"o}ding(2017)Steinegger and S{\"o}ding]{steinegger2017mmseqs2}
Martin Steinegger and Johannes S{\"o}ding.
\newblock Mmseqs2 enables sensitive protein sequence searching for the analysis of massive data sets.
\newblock \emph{Nature biotechnology}, 35\penalty0 (11):\penalty0 1026--1028, 2017.

\bibitem[Varadi et~al.(2024)Varadi, Bertoni, Magana, Paramval, Pidruchna, Radhakrishnan, Tsenkov, Nair, Mirdita, Yeo, et~al.]{varadi2024alphafold}
Mihaly Varadi, Damian Bertoni, Paulyna Magana, Urmila Paramval, Ivanna Pidruchna, Malarvizhi Radhakrishnan, Maxim Tsenkov, Sreenath Nair, Milot Mirdita, Jingi Yeo, et~al.
\newblock Alphafold protein structure database in 2024: providing structure coverage for over 214 million protein sequences.
\newblock \emph{Nucleic acids research}, 52\penalty0 (D1):\penalty0 D368--D375, 2024.

\bibitem[Wang et~al.(2025)Wang, Luo, Qin, Wang, Wan, Fang, Zhang, Gou, Su, Shen, et~al.]{wang20253dsmiles}
Jike Wang, Hao Luo, Rui Qin, Mingyang Wang, Xiaozhe Wan, Meijing Fang, Odin Zhang, Qiaolin Gou, Qun Su, Chao Shen, et~al.
\newblock 3dsmiles-gpt: 3d molecular pocket-based generation with token-only large language model.
\newblock \emph{Chemical Science}, 16\penalty0 (2):\penalty0 637--648, 2025.

\bibitem[Wu et~al.(2024)Wu, Xia, Deng, Liu, Zhang, Guo, Cui, Pei, Wu, Xie, et~al.]{wu2024tamgen}
Kehan Wu, Yingce Xia, Pan Deng, Renhe Liu, Yuan Zhang, Han Guo, Yumeng Cui, Qizhi Pei, Lijun Wu, Shufang Xie, et~al.
\newblock Tamgen: drug design with target-aware molecule generation through a chemical language model.
\newblock \emph{Nature Communications}, 15\penalty0 (1):\penalty0 9360, 2024.

\bibitem[Xia et~al.(2025)Xia, Jin, Xie, He, Cao, Luo, Liu, Wang, Liu, Chen, et~al.]{xia2025naturelm}
Yingce Xia, Peiran Jin, Shufang Xie, Liang He, Chuan Cao, Renqian Luo, Guoqing Liu, Yue Wang, Zequn Liu, Yuan-Jyue Chen, et~al.
\newblock Nature language model: Deciphering the language of nature for scientific discovery.
\newblock \emph{arXiv e-prints}, pp.\  arXiv--2502, 2025.

\bibitem[You et~al.(2018)You, Liu, Ying, Pande, and Leskovec]{you2018graph}
Jiaxuan You, Bowen Liu, Zhitao Ying, Vijay Pande, and Jure Leskovec.
\newblock Graph convolutional policy network for goal-directed molecular graph generation.
\newblock \emph{Advances in neural information processing systems}, 31, 2018.

\bibitem[Zhang et~al.(2024)Zhang, Gong, and Liu]{zhang2024rectified}
Daiheng Zhang, Chengyue Gong, and Qiang Liu.
\newblock Rectified flow for structure based drug design.
\newblock \emph{arXiv preprint arXiv:2412.01174}, 2024.

\bibitem[Zhang et~al.(2023)Zhang, Zhang, Jin, Zhang, Hu, Shen, Cao, Du, Kang, Deng, et~al.]{zhang2023resgen}
Odin Zhang, Jintu Zhang, Jieyu Jin, Xujun Zhang, RenLing Hu, Chao Shen, Hanqun Cao, Hongyan Du, Yu~Kang, Yafeng Deng, et~al.
\newblock Resgen is a pocket-aware 3d molecular generation model based on parallel multiscale modelling.
\newblock \emph{Nature Machine Intelligence}, 5\penalty0 (9):\penalty0 1020--1030, 2023.

\bibitem[Zholus et~al.(2025)Zholus, Kuznetsov, Schutski, Shayakhmetov, Polykovskiy, Chandar, and Zhavoronkov]{zholus2025bindgpt}
Artem Zholus, Maksim Kuznetsov, Roman Schutski, Rim Shayakhmetov, Daniil Polykovskiy, Sarath Chandar, and Alex Zhavoronkov.
\newblock Bindgpt: A scalable framework for 3d molecular design via language modeling and reinforcement learning.
\newblock In \emph{Proceedings of the AAAI Conference on Artificial Intelligence}, volume~39, pp.\  26083--26091, 2025.

\bibitem[Zhou et~al.(2023)Zhou, Gao, Ding, Zheng, Xu, Wei, Zhang, and Ke]{zhou2023uni}
Gengmo Zhou, Zhifeng Gao, Qiankun Ding, Hang Zheng, Hongteng Xu, Zhewei Wei, Linfeng Zhang, and Guolin Ke.
\newblock Uni-mol: A universal 3d molecular representation learning framework.
\newblock 2023.

\bibitem[Zhu et~al.(2025)Zhu, Shi, Bi, Jin, Liu, Zhang, Huang, Guo, Hu, Ju, et~al.]{zhu2025flexribbon}
Jianwei Zhu, Yu~Shi, Ran Bi, Peiran Jin, Chang Liu, Zhe Zhang, Haitao Huang, Zekun Guo, Pipi Hu, Fusong Ju, et~al.
\newblock Flexribbon: Joint sequence and structure pretraining for protein modeling.
\newblock \emph{bioRxiv}, pp.\  2025--10, 2025.

\bibitem[Ziegler et~al.(2019)Ziegler, Stiennon, Wu, Brown, Radford, Amodei, Christiano, and Irving]{ziegler2019fine}
Daniel~M Ziegler, Nisan Stiennon, Jeffrey Wu, Tom~B Brown, Alec Radford, Dario Amodei, Paul Christiano, and Geoffrey Irving.
\newblock Fine-tuning language models from human preferences.
\newblock \emph{arXiv preprint arXiv:1909.08593}, 2019.

\end{thebibliography}

\appendix
\section{Architecture}
\label{appendix:arch}


We provide detailed architectural descriptions for each component of \ours{}---the structure encoder, sequence generator, adapter, and VAE---with hyperparameters presented separately for each module.

\subsection{Structure Encoder}
\label{app:structure_encoder}

Our structure encoder follows the FlexRibbon framework~\citep{zhu2025flexribbon} but utilizes only its first two components: the sequence module and the structure module. The sequence module(encoder) maps proteins, molecules, and complexes into feature representations that support both intra-modal and cross-modal interactions. Meanwhile, the diffusion-based structure module (decoder) captures residue and atom distributions, yielding 3D coordinates and enriching representations with structural context. Detailed architectural hyperparameters are provided in Table~\ref{tab:encoder}.

\begin{table}[!htbp]
\centering
\caption{Hyperparameters of Structure Encoder.}
\begin{tabular}{lccc}
\toprule
Hyperparameters & Sequence Module & Structure Module \\
\midrule
Number of layers & 32 & 16 \\
Hidden size & 2048 & 2048 \\
FFN dimension & 8192 & 8192 \\
Attention heads & 32 & 32 \\
\bottomrule
\end{tabular}
\label{tab:encoder}
\end{table}

\paragraph{Sequence Module.}  
The sequence module separately processes protein sequences at the residue level and molecular graphs at the atom level, representing each as tokens.  A standard Transformer encoder is applied, where molecule tokens are augmented with learnable attention biases derived from their 2D topology, enabling the model to capture chemical connectivity. The resulting embeddings capture intra-protein, intra-molecule, and protein–molecule interactions, providing a comprehensive feature representation for subsequent modeling.

\paragraph{Structure Module.}  
The structure module is implemented as a Diffusion Transformer (DiT)~\citep{peebles2023scalable} that denoises the 3D coordinates of protein residues and molecular atoms. Coordinates are denoised under the conditioning of sequence-module representations, after being projected from 3D into a higher-dimensional latent space. Notably, ligand atoms require additional attention biases derived from bond connectivity. Through this design, this module refines noisy coordinates into chemically consistent structures, yielding enriched representations that couple spatial detail with sequence context for subsequent modeling.

\subsection{Sequence Generator} 
\label{app:sequence_gen}

We implement a reproduction of NatureLM-1B~\citep{xia2025naturelm}. The tokenizer is initialized from the LLaMA-3 vocabulary~\citep{dubey2024llama} (128,256 general-purpose tokens) and extended with a minimal set of domain-specific tokens: 26 for protein FASTA sequences, 1,401 for molecular SMILES strings, and four special markers ``$\langle{\rm mol}\rangle$'', ``$\langle{\rm /mol}\rangle$'', ``$\langle{\rm protein}\rangle$'', and ``$\langle{\rm /protein}\rangle$'' to indicate modality boundaries. Architectural hyperparameters are given in Table~3. The model is trained with a next-token prediction objective on both single-domain corpora (text, proteins, molecules) and cross-modal corpora (protein–text, molecule–text, protein–molecule–text), enabling it to retain general language modeling capacity while incorporating biomolecular semantics. The corresponding architectural hyperparameters are listed in Table~\ref{tab:generator}.

\begin{table}[!htbp]
\centering
\caption{Hyperparameters of Sequence Generator.}
\begin{tabular}{lccc}
\toprule
Hyperparameters & Value \\
\midrule
Vocabulary size & 129,687 \\
Number of layers & 16 \\
Hidden size & 2048 \\
FFN dimension & 5504 \\
Attention heads & 32 \\
\bottomrule
\end{tabular}
\label{tab:generator}
\end{table}

\subsection{Adapter and VAE}

\paragraph{Adapter} The adapter module provides a lightweight interface for injecting structural features into the language model. It adopts a gated MLP: input representations are processed by a gating projection and an up-projection, with the gated branch passing through a non-linear activation and combined element-wise with the up-projected features. A down-projection then maps the fused representation back to the hidden space, enabling efficient alignment with minimal additional parameters. Table~\ref{tab:adapter} reports the detailed architectural hyperparameters.

\begin{table}[!htbp]
\centering
\caption{Hyperparameters of Adapter.}
\begin{tabular}{lccc}
\toprule
Hyperparameters & Value \\
\midrule
Input dimension & 2048 \\
Intermediate dimension & 2048  \\
Output dimension & 2048 \\
\bottomrule
\end{tabular}
\label{tab:adapter}
\end{table}

\paragraph{VAE} The variational encoder maps complex representations into a latent Gaussian space using two MLPs that predict the mean and log-variance of the posterior. During training, it is only activated in Stage~B and Stage~C. The latent distribution of complex from structure encoder are injected as noise into the feature of the corresponding protein from structure encoder, thereby perturbing protein features and improving robustness. Architectural hyperparameters are summarized in Table\ref{tab:vae}.

\begin{table}[!htbp]
\centering
\caption{Hyperparameters of VAE.}
\begin{tabular}{lccc}
\toprule
Hyperparameters & Value \\
\midrule
Input dimension & 2048 \\
Latent dimension & 2048  \\
\bottomrule
\end{tabular}
\label{tab:vae}
\end{table}

\section{Implementation Details}



\paragraph{Structure Encoder Pre-training} 
The architecture of the structure encoder follows the Elucidated Diffusion Model (EDM)~\citep{karras2022elucidating}, a Transformer variant that integrates geometric information and has also been adopted in AlphaFold3~\citep{abramson2024accurate}. Following FlexRibbon~\citep{zhu2025flexribbon}, We pre-train the structure encoder on approximately 78M protein structures from AlphaFoldDB~\citep{varadi2024alphafold} and PDB~\citep{berman2000protein}, using a batch size of 4096 and a learning rate of $1{\times}10^{-4}$. 
Training is performed on 128 A100 GPUs for two weeks.

\paragraph{Alignment}  
Our alignment is implemented on a large-scale dataset of 1.1M instances (676K for proteins, 316K for molecules and 94K for complexes). The model is optimized with a learning rate of $1\times10^{-4}$, a batch size of 512, and 60K training steps, while keeping the backbone frozen and updating only the adapter parameters. Training was conducted on 32 A100 GPUs for 5 days.

\paragraph{Supervised Fine-tuning}  
For supervised fine-tuning, we use 100K examples from the CrossDocked2020 dataset. The model is optimized with a learning rate of $1\times10^{-5}$, a batch size of 128, and 15K training steps. The KL loss coefficient $\beta_{\text{vae}}$ is set to 0.1, and the VAE latent size is 2048. Training was performed on 8 A100 GPUs for approximately 30 hours.

\paragraph{Reinforcement Learning} 
For reinforcement learning with Direct Preference Optimization (DPO), we train on the $\mathcal{D}_{\text{DPO}}$ set consisting of 979 examples. The model is optimized with a learning rate of $5\times10^{-7}$ and a batch size of 8 for a single epoch, such that each sample is seen only once. The KL penalty coefficient $\beta_{\text{vae}}$ is set to 0.1 and the VAE latent size to 2048, identical to the SFT setting. Training is highly efficient and completes within 4 hours on 8 A100 GPUs with 112 vCPUs.

For online DPO, each protein pocket is used to generate 32 candidate molecules. Among valid generations, 5 are selected for docking, and the rewards described in the main text are used to construct best–worst preference pairs. The DPO loss employs $\beta_{\text{dpo}}=0.1$ to scale the advantage term in the preference objective. Sampling is performed with temperature $1.5$ and top-$p=0.95$ to encourage diversity, while a fused-ring penalty with weight $\lambda=0.5$ is applied to regularize chemical plausibility.

\paragraph{Inference}  
During inference, we sample molecules with temperature set to 1.5, a maximum generation length of 256 tokens, and top-$p=0.95$. For each protein pocket, at least 100 valid candidate molecules are generated to ensure sufficient diversity for downstream evaluation.

\section{Experiment Details and Supplementary Results}

\subsection{Experiment Details}

\begin{figure}[!htbp]
    \centering
    \includegraphics[width=0.9\linewidth]{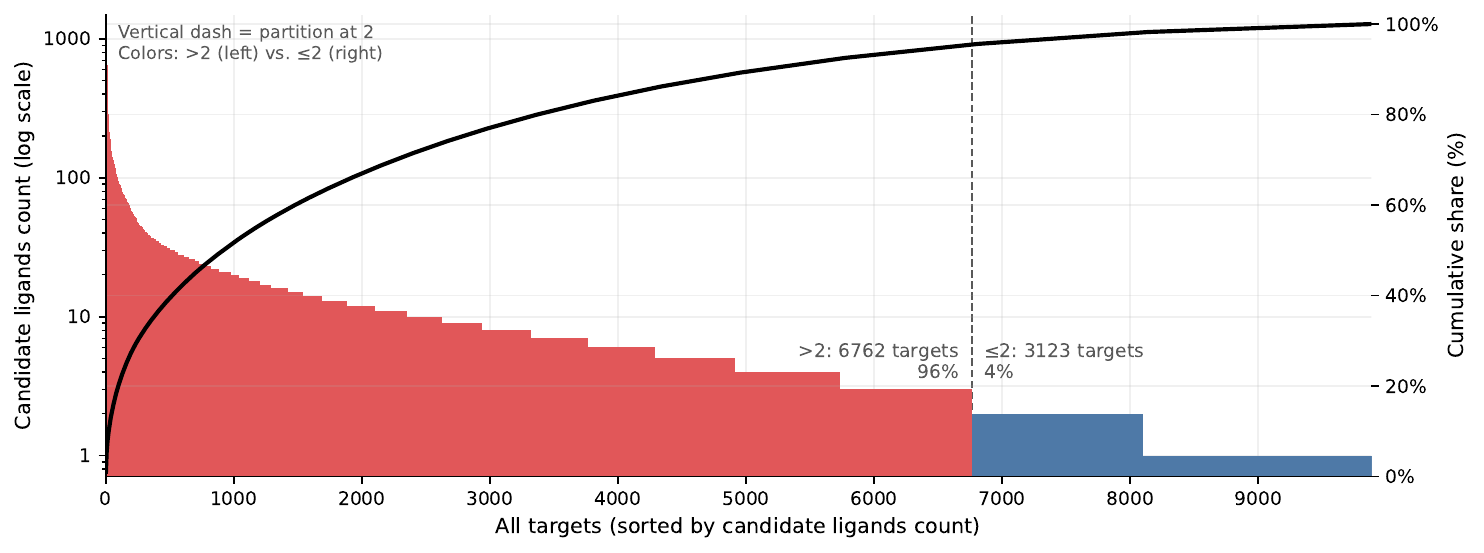}
    \caption{Distribution of candidate ligands per target in the CrossDocked2020 dataset. Targets are sorted by ligand count, with a dashed line marking the partition at 2 ligands, where the red and blue regions correspond to $\mathcal{D}_{\rm B}$ and $\mathcal{D}_{\rm C}$, respectively.}
    \label{fig:crossdock-data}
\end{figure}

\paragraph{Distribution Analysis of CrossDocked2020}
We visualize the distribution of candidate ligands per target in the CrossDocked2020 dataset, as shown in Figure~\ref{fig:crossdock-data}.

\paragraph{SA score} Note that the SA score is originally defined on a scale from 1 to 10~\citep{ertl2009estimation}, with lower values indicating greater synthesizability. Consistent with prior work on pocket-aware 3D drug design~\citep{guan20233d}, we apply a linear transformation, \(SA = (10 - SA_{\text{origin}}) / 9 \in [0, 1]\), so that higher values correspond to better synthesizability.

\paragraph{Generation Setup} In the structure-based drug design experiments, each baseline method generates no more than 100 molecules for a given protein pocket. In comparison, \ours{} produces exactly 100 unique molecules per pocket, enforcing a stricter evaluation protocol.

\paragraph{Docking Details}  
To evaluate docking, we convert generated SMILES strings into 3D molecular conformations. Molecules are first parsed with OpenBabel~\citep{o2011open} to obtain an initial structure, which is then processed with RDKit for conformer generation. We apply distance-geometry embedding with a fixed random seed, followed by MMFF optimization. If embedding fails, a 2D coordinate initialization is used as fallback. For docking, since our model does not generate binding poses directly, we use the center of the reference ligand as the pocket center. Docking is then performed with AutoDock Vina.

\subsection{Additional Results}  
In this subsection, we provide supplementary experimental results that could not be included in the main text. These include extended tables and figures for the main results, additional ablation studies, and further case study analyses.

\subsubsection{Main Results}

\begin{figure}[h]
    \centering
    \includegraphics[width=0.99\linewidth]{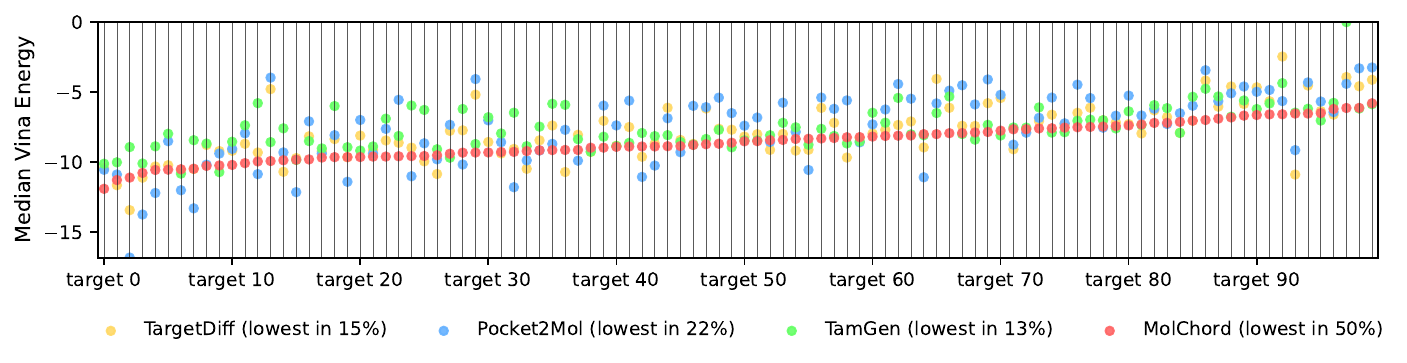}
    \caption{Median Vina energy for different generated molecules (TargetDiff, Pocket2Mol, TamGen, \ours{}) across 100 testing samples, sorted by the median Vina energy of molecules generated from \ours{}.}
    \label{median-vina-energy}
\end{figure}

\paragraph{Median Vina Energy} Figure~\ref{median-vina-energy} shows the median Vina energy of the proposed model, compared with TargetDiff, Pocket2Mol and TamGen, three representive methods in target-aware molecule generation. We observe that \ours{} surpasses these baseline models and generates molecules with the highest binding affinity for \(50\%\) of the protein targets in the test set. 

\paragraph{Fused Ring} The quantitative results of the fused-ring analysis is reported in Table~\ref{tab:fused-rings}

\begin{table}[!htbp]
\caption{Fused ring statistics for different generation methods.}
\centering
\begin{tabular}{lc}
\toprule
Method & Fused ring count \\
\midrule
TamGen       & 1.87 \\
Pocket2Mol  & 3.46 \\
ResGen      & 2.48 \\
TargetDiff    & 3.55 \\
\ours{} & 1.79 \\
\ours{}-RL & 1.75 \\
\bottomrule
\end{tabular}
\label{tab:fused-rings}
\end{table}

\paragraph{OOD Results}
The quantitative results of the OOD evaluation are reported in Table~\ref{tab:homology-ablation}. The full list of PDB IDs, comprising 40 homologous and 60 non-homologous cases, is provided below:

Homologous (40 cases): 4aaw, 4yhj, 14gs, 1fmc, 3g51, 2jjg, 4g3d, 5bur, 5q0k, 2azy, 5i0b, 1phk, 1djy, 5l1v, 4zfa, 4f1m, 4iwq, 5ngz, 1d7j, 4u5s, 3pdh, 1umd, 4pxz, 2gns, 1ai4, 5mma, 2cy0, 5d7n, 5mgl, 5aeh, 4xli, 3o96, 3hy9, 4bel, 4aua, 2f2c, 3chc, 1k9t, 1jn2, 4azf. 

Non-homologous (60 cases): 2z3h, 2v3r, 4rn0, 3daf, 1a2g, 5w2g, 3dzh, 1coy, 2rhy, 2pqw, 3gs6, 1r1h, 1dxo, 1gg5, 5b08, 4keu, 4q8b, 2rma, 3b6h, 2zen, 4p6p, 3u5y, 4tqr, 4lfu, 3jyh, 1l3l, 1e8h, 2e24, 2hcj, 3kc1, 4ja8, 4iiy, 3v4t, 3tym, 4d7o, 3ej8, 1rs9, 4kcq, 3w83, 2e6d, 4rv4, 1h36, 4gvd, 4tos, 4h3c, 4rlu, 3l3n, 5tjn, 5liu, 4qlk, 3nfb, 4m7t, 3u9f, 1h0i, 4z2g, 3af2, 3li4, 3pnm, 1afs, 2pc8.

\begin{table}[!htbp]
\centering
\caption{Comparison of average scores on homologous vs. non-homologous pockets, with $\Delta$ denoting their difference.}
\begin{tabular}{lccc}
\toprule
Method & Homologous (avg) & Non-homologous (avg) & $\Delta$ \\
\midrule
liGAN     & -6.31 & -5.98 & -0.33 \\
Pocket2Mol & -7.38 & -7.08 & -0.30 \\
TamGen    & -7.59 & -7.40 & -0.20 \\
\ours{}-RL  & -8.49 & -8.66 & \textbf{+0.17} \\
\bottomrule
\end{tabular}
\label{tab:homology-ablation}
\end{table}

\paragraph{Efficiency} \ours{} also demonstrates superior efficiency. On a single A100 GPU, it generates 100 compounds per target in $\sim$ 4s, while previous approaches such as Pocket2Mol, GraphBP, TargetDiff, DecompDiff, and ResGen typically require tens of seconds to several minutes. This highlights the practicality of \ours{} for large-scale molecular generation.

\subsubsection{Ablation Studies}

\paragraph{Effect of VAE}  
The effect of incorporating the VAE is shown in Table~\ref{crossdocked2020-ablation-vae}. We observe consistent gains across all evaluation metrics when the VAE is included, with particularly notable improvements in affinity-related measures. This can be attributed to the stochasticity introduced by the latent variables, which encourages broader exploration of the chemical space and enhances both molecular diversity and model robustness.

\begin{table}[!htbp]
\caption{The influence of VAE}
\label{crossdocked2020-ablation-vae}
\centering
\resizebox{\linewidth}{!}{
\begin{tabular}{lccccccccc}
\toprule
\multicolumn{1}{l}{Setting} & \multicolumn{1}{c}{Vina Dock ($\downarrow$)} & \multicolumn{1}{c}{High Affinity ($\uparrow$)} & \multicolumn{1}{c}{QED ($\uparrow$)} & \multicolumn{1}{c}{SA ($\uparrow$)} & \multicolumn{1}{c}{Diversity ($\uparrow$)} & \multicolumn{1}{c}{Success Rate ($\uparrow$)} \\ 
\midrule
\ours{} w/o VAE & -7.44 & 50.2\% & 0.55 & 0.76 & 0.75 & 29.5\% \\
\ours{} & -7.62 & 54.7\% & 0.56 & 0.77 & 0.76 & 33.2\% \\
\bottomrule
\end{tabular}
}
\end{table}

\section{Prompt}

All pre-training and fine-tuning tasks are formulated as text-augmented generation: structured entities (proteins, molecules, or complexes) are encoded into feature vectors by the structure encoder and injected into reserved slots of the language model’s input embeddings. 
Placeholders marked as ``\texttt{$\langle{\rm 3d\; protein}\rangle$}'', ``\texttt{$\langle{\rm 3d\; molecule}\rangle$}'', or ``\texttt{$\langle{\rm 3d\; complex}\rangle$}'' are routed to the structure encoder rather than tokenized, and their features replace the corresponding placeholder tokens in the prompt embedding.

\subsection{Protein Alignment Prompts}
All protein prompts follow a unified template, where the structured placeholder \texttt{$\langle{\rm 3d\; protein}\rangle$} is encoded by the structure encoder, \texttt{[FASTA]} specifies the protein sequence, and \texttt{[description]} provides the textual description.
For example:  

\texttt{Compose a summary of the protein $\langle{\rm 3d\; protein}\rangle$, which is also [FASTA]. [description]}

To improve robustness, we paraphrase the instruction into multiple variants while keeping the same format (see Table~\ref{tab:protein-prompts} for the full list).

\subsection{Molecule Alignment Prompts.}   
Similar to proteins, all molecule prompts follow a unified template, where the structured placeholder \texttt{$\langle{\rm 3d\;molecule}\rangle$} is encoded by the structure encoder, \texttt{[SMILES]} represents the molecule SMILES string, and \texttt{[description]} provides the textual description.  
For example:  

\texttt{Give a breakdown of the molecule $\langle{\rm 3d\; molecule}\rangle$, which is also [SMILES]. [description]}

To improve robustness, we paraphrase the instruction into multiple variants while keeping the same format (see Table~\ref{tab:molecule-prompts} for the full list).

\subsection{Complex Alignment Prompts.}  
For protein–ligand complexes, all prompts follow a unified template: an instruction applied to the structured placeholder $\langle{\rm 3d\; complex}\rangle$, which internally consists of a protein sequence (FASTA) and a molecule (SMILES).  
For example:  

\texttt{The protein-ligand complex $\langle{\rm 3d\; complex}\rangle$ consists of protein [FASTA] and molecule [SMILES].}

Here, \texttt{[FASTA]} and \texttt{[SMILES]} denote the textual placeholders for the protein sequence and molecular string, respectively.  
To improve robustness, we paraphrase the instruction into multiple variants while keeping the same format (see Table~\ref{tab:complex-prompts} for the full list).

\subsection{Structure-Based Drug Design}  
In the structure-based setting, we design prompts that condition ligand generation on protein binding pockets. Each prompt follows a unified template: an instruction followed by the structured placeholder \texttt{$\langle{\rm 3d\; pocket}\rangle$}, which is encoded by the structure encoder.  
For example:

\texttt{Generate a compound based on the pocket $\langle{\rm 3d\; pocket}\rangle$.}

Notably, these prompts are used in Stage B, where only the pocket features are concatenated with text embeddings. For the ablation in Stage A, the same templates are used with both the placeholder and the keyword ``pocket'' replaced by ``protein,'' ensuring that generation is conditioned on full protein features rather than pocket features. The full list of paraphrased SBDD prompts is provided in Table~\ref{tab:sbdd-prompts}.

\begin{table}[!tbp]
\centering
\caption{Full list of paraphrased protein prompts, where \texttt{$\langle{\rm 3d\; protein}\rangle$} is encoded by the structure encoder, \texttt{[FASTA]} specifies the protein sequence, and \texttt{[description]} provides the textual description.
}
\scriptsize
\begin{tabular}{p{0.96\linewidth}}
Give a breakdown of the protein sequence $\langle{\rm 3d\; protein}\rangle$, which is also [FASTA]. [description] \\
Give a breakdown of the FASTA sequence $\langle{\rm 3d\; protein}\rangle$, which is also [FASTA]. [description] \\
Establish an interpretation of the protein sequence $\langle{\rm 3d\; protein}\rangle$, which is also [FASTA]. [description] \\
Establish an interpretation of the FASTA sequenceCreate a representation of the protein sequence's description $\langle{\rm 3d\; protein}\rangle$, which is also [FASTA]. [description] \\
Create a representation of the FASTA sequence's description $\langle{\rm 3d\; protein}\rangle$, which is also [FASTA]. [description] \\
Formulate an explanation of the protein sequence $\langle{\rm 3d\; protein}\rangle$, which is also [FASTA]. [description] \\
Formulate an explanation of the FASTA sequence $\langle{\rm 3d\; protein}\rangle$, which is also [FASTA]. [description] \\
Construct a depiction of the protein sequence $\langle{\rm 3d\; protein}\rangle$, which is also [FASTA]. [description] \\
Construct a depiction of the FASTA sequence $\langle{\rm 3d\; protein}\rangle$, which is also [FASTA]. [description] \\
Form a presentation of the protein sequence $\langle{\rm 3d\; protein}\rangle$, which is also [FASTA]. [description] \\
Form a presentation of the FASTA sequence $\langle{\rm 3d\; protein}\rangle$, which is also [FASTA]. [description] \\
Develop a narrative for the protein sequence $\langle{\rm 3d\; protein}\rangle$, which is also [FASTA]. [description] \\
Develop a narrative for the FASTA sequence $\langle{\rm 3d\; protein}\rangle$, which is also [FASTA]. [description] \\
Prepare a profile of the protein sequence $\langle{\rm 3d\; protein}\rangle$, which is also [FASTA]. [description] \\
Prepare a profile of the FASTA sequence $\langle{\rm 3d\; protein}\rangle$, which is also [FASTA]. [description] \\
Illustrate the characteristics of the protein sequence $\langle{\rm 3d\; protein}\rangle$, which is also [FASTA]. [description] \\
Illustrate the characteristics of the FASTA sequence $\langle{\rm 3d\; protein}\rangle$, which is also [FASTA]. [description] \\
Present a report on the protein sequence $\langle{\rm 3d\; protein}\rangle$, which is also [FASTA]. [description] \\
Present a report on the FASTA sequence $\langle{\rm 3d\; protein}\rangle$, which is also [FASTA]. [description] \\
Generate the description of the protein sequence $\langle{\rm 3d\; protein}\rangle$, which is also [FASTA]. [description] \\
Generate the description of the FASTA sequence $\langle{\rm 3d\; protein}\rangle$, which is also [FASTA]. [description] \\
Offer an analysis of the protein sequence $\langle{\rm 3d\; protein}\rangle$, which is also [FASTA]. [description] \\
Offer an analysis of the FASTA sequence $\langle{\rm 3d\; protein}\rangle$, which is also [FASTA]. [description] \\
Render an explication of the protein sequence $\langle{\rm 3d\; protein}\rangle$, which is also [FASTA]. [description] \\
Render an explication of the FASTA sequence $\langle{\rm 3d\; protein}\rangle$, which is also [FASTA]. [description] \\
Set forth an elucidation of the protein sequence $\langle{\rm 3d\; protein}\rangle$, which is also [FASTA]. [description] \\
Set forth an elucidation of the FASTA sequence $\langle{\rm 3d\; protein}\rangle$, which is also [FASTA]. [description] \\
Compose a summary of the protein sequence $\langle{\rm 3d\; protein}\rangle$, which is also [FASTA]. [description] \\
Compose a summary of the FASTA sequence $\langle{\rm 3d\; protein}\rangle$, which is also [FASTA]. [description] \\
Draw up a delineation of the protein sequence $\langle{\rm 3d\; protein}\rangle$, which is also [FASTA]. [description] \\
Draw up a delineation of the FASTA sequence $\langle{\rm 3d\; protein}\rangle$, which is also [FASTA]. [description] \\
Assemble a sketch of the protein sequence $\langle{\rm 3d\; protein}\rangle$, which is also [FASTA]. [description] \\
Assemble a sketch of the FASTA sequence $\langle{\rm 3d\; protein}\rangle$, which is also [FASTA]. [description] \\
Provide an overview of the protein sequence $\langle{\rm 3d\; protein}\rangle$, which is also [FASTA]. [description] \\
Provide an overview of the FASTA sequence $\langle{\rm 3d\; protein}\rangle$, which is also [FASTA]. [description] \\
Craft an outline of the protein sequence $\langle{\rm 3d\; protein}\rangle$, which is also [FASTA]. [description] \\
Craft an outline of the FASTA sequence $\langle{\rm 3d\; protein}\rangle$, which is also [FASTA]. [description] \\
Produce a detailed account of the protein sequence $\langle{\rm 3d\; protein}\rangle$, which is also [FASTA]. [description] \\
Produce a detailed account of the FASTA sequence $\langle{\rm 3d\; protein}\rangle$, which is also [FASTA]. [description] \\
Build a portrayal of the protein sequence $\langle{\rm 3d\; protein}\rangle$, which is also [FASTA]. [description] \\
Build a portrayal of the FASTA sequence $\langle{\rm 3d\; protein}\rangle$, which is also [FASTA]. [description] \\
\end{tabular}
\label{tab:protein-prompts}
\end{table}

\begin{table}[!tbp]
\centering
\caption{Full list of paraphrased molecule prompts, where \texttt{$\langle{\rm 3d\; molecule}\rangle$} is encoded by the structure encoder, \texttt{[SMILES]} specifies the molecular representation string, and \texttt{[description]} denotes the textual description.}
\scriptsize

\begin{tabular}{p{0.96\linewidth}}
Give a breakdown of the chemical compound $\langle{\rm 3d\; molecule}\rangle$, which is also [SMILES]. [description] \\
Give a breakdown of the molecule $\langle{\rm 3d\; molecule}\rangle$, which is also [SMILES]. [description] \\
Give a breakdown of the SMILES string $\langle{\rm 3d\; molecule}\rangle$, which is also [SMILES]. [description] \\
Establish an interpretation of the chemical compound $\langle{\rm 3d\; molecule}\rangle$, which is also [SMILES]. [description] \\
Establish an interpretation of the molecule $\langle{\rm 3d\; molecule}\rangle$, which is also [SMILES]. [description] \\
Establish an interpretation of the SMILES stringCreate a representation of the chemical compound's description $\langle{\rm 3d\; molecule}\rangle$, which is also [SMILES]. [description] \\
Create a representation of the molecule's description $\langle{\rm 3d\; molecule}\rangle$, which is also [SMILES]. [description] \\
Create a representation of the SMILES string's description $\langle{\rm 3d\; molecule}\rangle$, which is also [SMILES]. [description] \\
Formulate an explanation of the chemical compound $\langle{\rm 3d\; molecule}\rangle$, which is also [SMILES]. [description] \\
Formulate an explanation of the molecule $\langle{\rm 3d\; molecule}\rangle$, which is also [SMILES]. [description] \\
Formulate an explanation of the SMILES string $\langle{\rm 3d\; molecule}\rangle$, which is also [SMILES]. [description] \\
Construct a depiction of the chemical compound $\langle{\rm 3d\; molecule}\rangle$, which is also [SMILES]. [description] \\
Construct a depiction of the molecule $\langle{\rm 3d\; molecule}\rangle$, which is also [SMILES]. [description] \\
Construct a depiction of the SMILES string $\langle{\rm 3d\; molecule}\rangle$, which is also [SMILES]. [description] \\
Form a presentation of the chemical compound $\langle{\rm 3d\; molecule}\rangle$, which is also [SMILES]. [description] \\
Form a presentation of the molecule $\langle{\rm 3d\; molecule}\rangle$, which is also [SMILES]. [description] \\
Form a presentation of the SMILES string $\langle{\rm 3d\; molecule}\rangle$, which is also [SMILES]. [description] \\
Develop a narrative for the chemical compound $\langle{\rm 3d\; molecule}\rangle$, which is also [SMILES]. [description] \\
Develop a narrative for the molecule $\langle{\rm 3d\; molecule}\rangle$, which is also [SMILES]. [description] \\
Develop a narrative for the SMILES string $\langle{\rm 3d\; molecule}\rangle$, which is also [SMILES]. [description] \\
Prepare a profile of the chemical compound $\langle{\rm 3d\; molecule}\rangle$, which is also [SMILES]. [description] \\
Prepare a profile of the molecule $\langle{\rm 3d\; molecule}\rangle$, which is also [SMILES]. [description] \\
Prepare a profile of the SMILES string $\langle{\rm 3d\; molecule}\rangle$, which is also [SMILES]. [description] \\
Illustrate the characteristics of the chemical compound $\langle{\rm 3d\; molecule}\rangle$, which is also [SMILES]. [description] \\
Illustrate the characteristics of the molecule $\langle{\rm 3d\; molecule}\rangle$, which is also [SMILES]. [description] \\
Illustrate the characteristics of the SMILES string $\langle{\rm 3d\; molecule}\rangle$, which is also [SMILES]. [description] \\
Present a report on the chemical compound $\langle{\rm 3d\; molecule}\rangle$, which is also [SMILES]. [description] \\
Present a report on the molecule $\langle{\rm 3d\; molecule}\rangle$, which is also [SMILES]. [description] \\
Present a report on the SMILES string $\langle{\rm 3d\; molecule}\rangle$, which is also [SMILES]. [description] \\
Generate the description of the chemical compound $\langle{\rm 3d\; molecule}\rangle$, which is also [SMILES]. [description] \\
Generate the description of the molecule $\langle{\rm 3d\; molecule}\rangle$, which is also [SMILES]. [description] \\
Generate the description of the SMILES string $\langle{\rm 3d\; molecule}\rangle$, which is also [SMILES]. [description] \\
Offer an analysis of the chemical compound $\langle{\rm 3d\; molecule}\rangle$, which is also [SMILES]. [description] \\
Offer an analysis of the molecule $\langle{\rm 3d\; molecule}\rangle$, which is also [SMILES]. [description] \\
Offer an analysis of the SMILES string $\langle{\rm 3d\; molecule}\rangle$, which is also [SMILES]. [description] \\
Render an explication of the chemical compound $\langle{\rm 3d\; molecule}\rangle$, which is also [SMILES]. [description] \\
Render an explication of the molecule $\langle{\rm 3d\; molecule}\rangle$, which is also [SMILES]. [description] \\
Render an explication of the SMILES string $\langle{\rm 3d\; molecule}\rangle$, which is also [SMILES]. [description] \\
Set forth an elucidation of the chemical compound $\langle{\rm 3d\; molecule}\rangle$, which is also [SMILES]. [description] \\
Set forth an elucidation of the molecule $\langle{\rm 3d\; molecule}\rangle$, which is also [SMILES]. [description] \\
Set forth an elucidation of the SMILES string $\langle{\rm 3d\; molecule}\rangle$, which is also [SMILES]. [description] \\
Compose a summary of the chemical compound $\langle{\rm 3d\; molecule}\rangle$, which is also [SMILES]. [description] \\
Compose a summary of the molecule $\langle{\rm 3d\; molecule}\rangle$, which is also [SMILES]. [description] \\
Compose a summary of the SMILES string $\langle{\rm 3d\; molecule}\rangle$, which is also [SMILES]. [description] \\
Draw up a delineation of the chemical compound $\langle{\rm 3d\; molecule}\rangle$, which is also [SMILES]. [description] \\
Draw up a delineation of the molecule $\langle{\rm 3d\; molecule}\rangle$, which is also [SMILES]. [description] \\
Draw up a delineation of the SMILES string $\langle{\rm 3d\; molecule}\rangle$, which is also [SMILES]. [description] \\
Assemble a sketch of the chemical compound $\langle{\rm 3d\; molecule}\rangle$, which is also [SMILES]. [description] \\
Assemble a sketch of the molecule $\langle{\rm 3d\; molecule}\rangle$, which is also [SMILES]. [description] \\
Assemble a sketch of the SMILES string $\langle{\rm 3d\; molecule}\rangle$, which is also [SMILES]. [description] \\
Provide an overview of the chemical compound $\langle{\rm 3d\; molecule}\rangle$, which is also [SMILES]. [description] \\
Provide an overview of the molecule $\langle{\rm 3d\; molecule}\rangle$, which is also [SMILES]. [description] \\
Provide an overview of the SMILES string $\langle{\rm 3d\; molecule}\rangle$, which is also [SMILES]. [description] \\
Craft an outline of the chemical compound $\langle{\rm 3d\; molecule}\rangle$, which is also [SMILES]. [description] \\
Craft an outline of the molecule $\langle{\rm 3d\; molecule}\rangle$, which is also [SMILES]. [description] \\
Craft an outline of the SMILES string $\langle{\rm 3d\; molecule}\rangle$, which is also [SMILES]. [description] \\
Produce a detailed account of the chemical compound $\langle{\rm 3d\; molecule}\rangle$, which is also [SMILES]. [description] \\
Produce a detailed account of the molecule $\langle{\rm 3d\; molecule}\rangle$, which is also [SMILES]. [description] \\
Produce a detailed account of the SMILES string $\langle{\rm 3d\; molecule}\rangle$, which is also [SMILES]. [description] \\
Build a portrayal of the chemical compound $\langle{\rm 3d\; molecule}\rangle$, which is also [SMILES]. [description] \\
Build a portrayal of the molecule $\langle{\rm 3d\; molecule}\rangle$, which is also [SMILES]. [description] \\
Build a portrayal of the SMILES string $\langle{\rm 3d\; molecule}\rangle$, which is also [SMILES]. [description] \\
\end{tabular}
\label{tab:molecule-prompts}
\end{table}

\begin{table}[!tbp]
\centering
\caption{Full list of paraphrased complex prompts, where \texttt{$\langle{\rm 3d\; complex}\rangle$} is a structured placeholder rather than a textual input, with \texttt{[FASTA]} specifying the protein sequence and \texttt{[SMILES]} specifying the molecular representation string corresponding to the ligand component.
}
\scriptsize
\begin{tabular}{p{0.96\linewidth}}
The protein-ligand complex $\langle{\rm 3d\; complex}\rangle$ consists of protein [FASTA] and molecule [SMILES]. \\
The following protein-ligand pair $\langle{\rm 3d\; complex}\rangle$ contains a protein [FASTA] and a compound [SMILES]. \\
This complex $\langle{\rm 3d\; complex}\rangle$ is formed by protein [FASTA] and ligand [SMILES]. \\
The complex $\langle{\rm 3d\; complex}\rangle$ involves a protein sequence [FASTA] and a chemical compound [SMILES]. \\
Here is a protein-ligand complex $\langle{\rm 3d\; complex}\rangle$ comprising [FASTA] and [SMILES]. \\
The input complex $\langle{\rm 3d\; complex}\rangle$ includes protein [FASTA] and chemical compound [SMILES]. \\
The structure $\langle{\rm 3d\; complex}\rangle$ represents a binding between protein [FASTA] and molecule [SMILES]. \\
The biomolecular pair $\langle{\rm 3d\; complex}\rangle$ consists of protein [FASTA] and SMILES representation [SMILES]. \\
In this complex $\langle{\rm 3d\; complex}\rangle$, a protein [FASTA] interacts with a compound [SMILES]. \\
This protein-ligand pair $\langle{\rm 3d\; complex}\rangle$ includes a protein structure [FASTA] and a molecular graph [SMILES]. \\
The following complex $\langle{\rm 3d\; complex}\rangle$ illustrates a molecular interaction between [FASTA] and [SMILES]. \\
This protein-ligand complex $\langle{\rm 3d\; complex}\rangle$ is composed of protein [FASTA] and chemical entity [SMILES]. \\
In the provided complex $\langle{\rm 3d\; complex}\rangle$, the protein [FASTA] is paired with ligand [SMILES]. \\
The example complex $\langle{\rm 3d\; complex}\rangle$ is constructed from a protein [FASTA] and molecule [SMILES]. \\
$\langle{\rm 3d\; complex}\rangle$ is a protein-ligand pair composed of sequence [FASTA] and SMILES [SMILES]. \\
In this molecular protein-ligand complex $\langle{\rm 3d\; complex}\rangle$, we observe the interaction between [FASTA] and [SMILES]. \\
The complex $\langle{\rm 3d\; complex}\rangle$ showcases a biochemical pair of [FASTA] and [SMILES]. \\
The protein-ligand complex $\langle{\rm 3d\; complex}\rangle$ links [FASTA] with [SMILES]. \\
The pair $\langle{\rm 3d\; complex}\rangle$ includes a protein [FASTA] and its corresponding ligand [SMILES]. \\
The following structure $\langle{\rm 3d\; complex}\rangle$ shows a protein-ligand interaction between [FASTA] and [SMILES]. \\
The complex $\langle{\rm 3d\; complex}\rangle$ represents the molecular interaction of sequence [FASTA] and structure [SMILES]. \\
This biomolecular structure $\langle{\rm 3d\; complex}\rangle$ is composed of [FASTA] and [SMILES]. \\
$\langle{\rm 3d\; complex}\rangle$ depicts a protein-ligand binding between protein [FASTA] and molecule [SMILES]. \\
In $\langle{\rm 3d\; complex}\rangle$, the protein target [FASTA] is complexed with small molecule [SMILES]. \\
The given molecular complex $\langle{\rm 3d\; complex}\rangle$ combines protein [FASTA] and ligand [SMILES]. \\
\end{tabular}
\label{tab:complex-prompts}
\end{table}

\begin{table}[!t]
\centering
\caption{Full list of paraphrased SBDD prompts in Stage~B, where \texttt{$\langle{\rm3d\; pocket}\rangle$} denotes the protein pocket. For the Stage~A ablation, both the keyword and placeholder ``pocket'' are replaced by ``protein''.}
\scriptsize
\begin{tabular}{p{0.96\linewidth}}
Generate a compound based on the pocket $\langle{\rm 3d\; pocket}\rangle$. \\
Innovate a compound with the pocket $\langle{\rm 3d\; pocket}\rangle$ as a foundation. \\
Assemble a compound in relation to the pocket $\langle{\rm 3d\; pocket}\rangle$. \\
Create a compound influenced by the pocket $\langle{\rm 3d\; pocket}\rangle$. \\
Construct a compound reflecting the essence of the pocket $\langle{\rm 3d\; pocket}\rangle$. \\
Prepare a compound derived from the principles of the pocket $\langle{\rm 3d\; pocket}\rangle$. \\
Innovate a compound in the spirit of the pocket $\langle{\rm 3d\; pocket}\rangle$. \\
Develop a compound that matches the pocket $\langle{\rm 3d\; pocket}\rangle$. \\
Synthesize a compound derived from the pocket $\langle{\rm 3d\; pocket}\rangle$. \\
Manufacture a compound using the pocket $\langle{\rm 3d\; pocket}\rangle$ as a basis. \\
Create a compound that corresponds to the pocket $\langle{\rm 3d\; pocket}\rangle$. \\
Generate a compound that aligns with the pocket $\langle{\rm 3d\; pocket}\rangle$. \\
Synthesize a compound according to the pocket $\langle{\rm 3d\; pocket}\rangle$. \\
Craft a compound in the likeness of the pocket $\langle{\rm 3d\; pocket}\rangle$. \\
Assemble a compound inspired by the essence of the pocket $\langle{\rm 3d\; pocket}\rangle$. \\
Formulate a compound in accordance with the pocket $\langle{\rm 3d\; pocket}\rangle$. \\
Fabricate a compound that adheres to the pocket $\langle{\rm 3d\; pocket}\rangle$. \\
Engineer a compound anchored in the pocket $\langle{\rm 3d\; pocket}\rangle$. \\
Craft a compound that embodies the pocket $\langle{\rm 3d\; pocket}\rangle$. \\
Cultivate a compound with the pocket $\langle{\rm 3d\; pocket}\rangle$ in mind. \\
Design a compound that conforms to the pocket $\langle{\rm 3d\; pocket}\rangle$. \\
Formulate a compound that is influenced by the pocket $\langle{\rm 3d\; pocket}\rangle$. \\
Produce a compound guided by the pocket $\langle{\rm 3d\; pocket}\rangle$. \\
Construct a compound modeled on the pocket $\langle{\rm 3d\; pocket}\rangle$. \\
Design a compound with reference to the pocket $\langle{\rm 3d\; pocket}\rangle$. \\
Generate a compound reflecting the attributes of the pocket $\langle{\rm 3d\; pocket}\rangle$. \\
Produce a compound that incorporates the pocket $\langle{\rm 3d\; pocket}\rangle$. \\
Formulate a compound that mirrors the pocket $\langle{\rm 3d\; pocket}\rangle$. \\
Fabricate a compound utilizing the pocket $\langle{\rm 3d\; pocket}\rangle$. \\
Develop a compound that is rooted in the pocket $\langle{\rm 3d\; pocket}\rangle$. \\
Create a compound that is consistent with the pocket $\langle{\rm 3d\; pocket}\rangle$. \\
Assemble a compound taking the pocket $\langle{\rm 3d\; pocket}\rangle$ into account. \\
Derive a compound from the characteristics of the pocket $\langle{\rm 3d\; pocket}\rangle$. \\
Produce a compound based on the criteria of the pocket $\langle{\rm 3d\; pocket}\rangle$. \\
Compose a compound centered around the pocket $\langle{\rm 3d\; pocket}\rangle$. \\
Fashion a compound in response to the pocket $\langle{\rm 3d\; pocket}\rangle$. \\
Invent a compound informed by the pocket $\langle{\rm 3d\; pocket}\rangle$. \\
Devise a compound inspired by the pocket $\langle{\rm 3d\; pocket}\rangle$. \\
Construct a compound that reflects the pocket $\langle{\rm 3d\; pocket}\rangle$. \\
Design a compound following the pocket $\langle{\rm 3d\; pocket}\rangle$. \\
Develop a compound referencing the pocket $\langle{\rm 3d\; pocket}\rangle$. \\
\end{tabular}
\label{tab:sbdd-prompts}
\end{table}

\section{Usage of LLM}  
We employed large language models (GPT-5 and GPT-4o) as auxiliary tools during paper writing. Their usage was confined to non-technical writing support, including grammar checking, stylistic adjustments, and improvements in clarity and fluency. All technical ideas, dataset construction, experimental design, and result analysis originate solely from the authors. The use of LLMs did not contribute to the scientific content of this work and served only to facilitate more fluent and polished writing.

\end{document}